\RequirePackage{fix-cm}
\documentclass[twocolumn]{svjour3}          
\smartqed  
\newlength\savedwidth
\newcommand{\whline}{\noalign{\global\savedwidth\arrayrulewidth
                            \global\arrayrulewidth 1.5pt}%
                   \hline
                   \noalign{\global\arrayrulewidth\savedwidth}}

\newcommand{\swhline}{\noalign{\global\savedwidth\arrayrulewidth
\global\arrayrulewidth 1.2pt}%
\hline
\noalign{\global\arrayrulewidth\savedwidth}}

\newcommand{\changed}[1]{{\color{black}{#1}}}

\usepackage[floatrowsep=none]{floatrow}
\usepackage{bm}
\usepackage{enumitem}
\usepackage{floatrow}
\usepackage{multirow}
\usepackage{subfigure}
\usepackage[ruled]{algorithm2e}
\usepackage{graphicx}
\usepackage{amssymb}
\usepackage{amsmath} 
\usepackage{tabularx}
\usepackage{colortbl}
\usepackage[toc,page]{appendix}

\usepackage{times}
\usepackage{rotating}
\usepackage{bbm}
\usepackage{color}
\usepackage{threeparttable}
\usepackage{rotating}
\usepackage{makecell}

\usepackage{float}
\floatstyle{plaintop}
\restylefloat{table}

\usepackage[toc,page]{appendix}
\usepackage[round]{natbib}

\newcommand{\eddy}[1]{{\color{black}#1}}
\newcommand{\sgg}[1]{{\color{black}#1}}

\definecolor{Gray}{gray}{0.9}
\begin{document}

\title{Person Re-Identification in Identity Regression Space
}


\author{Hanxiao Wang         \and
        Xiatian Zhu \and
        Shaogang Gong \and
        Tao Xiang
}


\institute{
			Hanxiao Wang is with Electrical and Computer Engineering
			Department,
			Boston University,
			Boston MA 02215, USA.
			\email{hxw@bu.edu}.
			\\
			Xiatian Zhu is with Vision Semantics Limited, 
			London E1 4NS, UK.
			\email{eddy@visionsemantics.com}.
			\\            
            Shaogang Gong and Tao Xiang are with
            School of Electronic Engineering and
            Computer Science,
            Queen Mary University of London, London E1 4NS, UK.
            \email{\{s.gong, t.xiang\}@qmul.ac.uk}.
           %
}

\date{Received: date / Accepted: date}

\maketitle

\begin{abstract}
\changed{Most existing person re-identification (re-id) methods are unsuitable
for real-world deployment due to two reasons: 
{\em Unscalability to large population size}, and
{\em Inadaptability over time}.
In this work, we present a unified solution to address
both problems. Specifically, we propose to construct an
Identity Regression Space (IRS) based on embedding different training person identities
(classes) and formulate re-id as a
regression problem solved by identity regression in the IRS. 
The IRS approach is characterised by
a closed-form solution with high learning efficiency and
an inherent incremental learning capability with human-in-the-loop. 
Extensive experiments on four benchmarking datasets (VIPeR,
CUHK01, CUHK03 and Market-1501) show that 
the IRS model not only outperforms 
state-of-the-art re-id methods, but also is more scalable to large re-id population size
by rapidly updating model and 
actively selecting informative samples with reduced human labelling effort.}

\end{abstract}

\section{Introduction}
\label{sec:intro}

\begin{figure} 
	\centering
	\subfigure[Significant person appearance change across camera views.]{
		\includegraphics[width=1\linewidth]{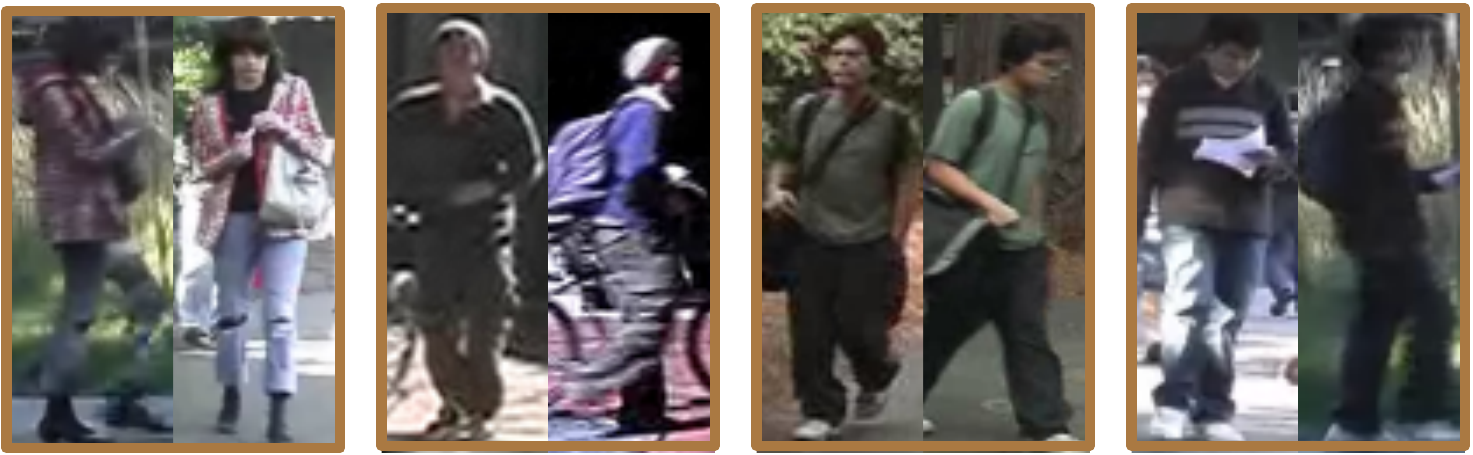}
	}
	\subfigure[High visual similarity among different people.]{
		\includegraphics[width=1\linewidth]{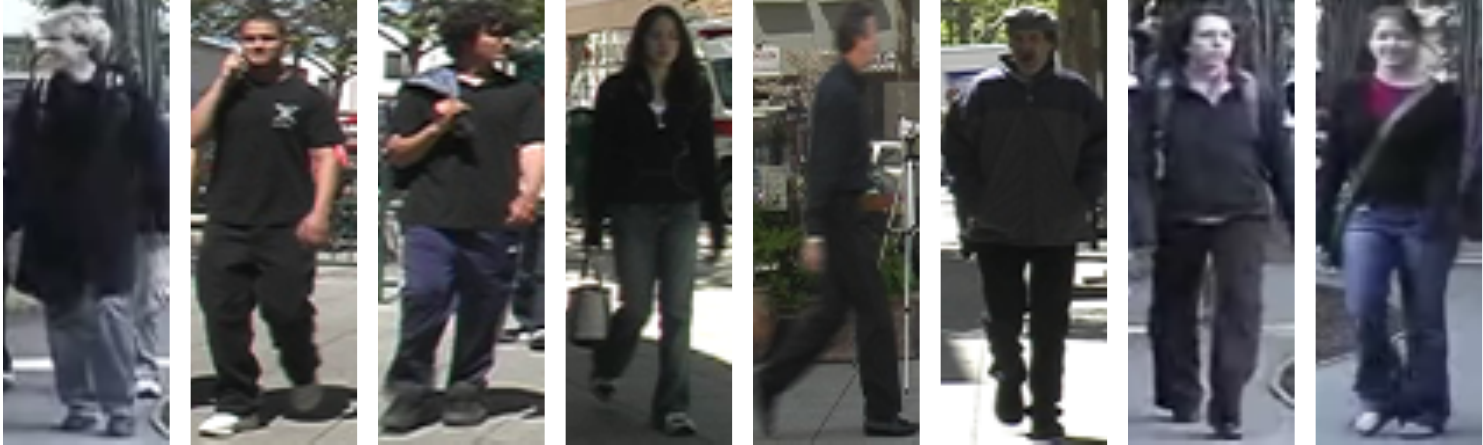}
	}
	\vskip -0.5cm
	\caption{
		Illustration of person re-identification challenges.
	}
	\label{fig:problem}
\end{figure}

\changed{Person re-identification (re-id) aims to match
identity classes of person images captured under 
non-overlapping camera views
\citep{gong2014person}.
It is inherently challenging
due to significant cross-view appearance changes (Fig. \ref{fig:problem}(a))
and high visual similarity among different people (Fig. \ref{fig:problem}(b)).
Most existing re-id methods focus on 
designing identity discriminative features and matching models 
for reducing intra-person appearance disparity whilst increasing inter-person appearance individuality.
This is often formulated as a
supervised learning problem through
classification
\citep{KISSME_CVPR12,liao2015person},
pairwise verification 
\citep{Li_DeepReID_2014b,shi2016embedding},
triplet ranking \citep{PRDC,wang2016pami},
or a combination thereof
\citep{wangjoint}.
While achieving ever-increasing re-id performance on
benchmarking datasets \citep{zheng2016person,karanam2016comprehensive},
these methods are restricted in scaling up to real-world deployments 
due to two fundamental limitations:}

\changed{{\bf (I)} {\em Small Sample Size}: 
The labelled training population is often small
(e.g. hundreds of persons each with a few images)
and much smaller (e.g. $<$$\frac{1}{10}$) 
than typical feature dimensions.
This is because collecting cross-view matched image pairs 
from different locations is not only tedious but also difficult.
The lack of training samples is known as the Small
Sample Size (SSS) problem \citep{chen2000new}, 
which may cause singular intra-class and poor
inter-class scatter matrices.
Given that metric learning re-id methods
aim to minimise the within-class
(intra-person) variance whilst maximising the inter-class (inter-person) variance,
the SSS problem is therefore likely to make the solutions suboptimal.

{\bf (II)} {\em Inadaptability}: 
Existing re-id methods often adopt off-line batch-wise
model learning with the need for sufficiently large sized training data collected via
a time consuming manual labelling process. 
This {\em first-labelling-then-training} scheme  
is not scalable to real-world applications that require
deployments at many previously unseen surveillance locations 
with little or no
labelled data in advance.
Also, real-world label collection is more incremental,
i.e. additional label data are sequentially available for model
update over time.
It is hence desirable for a re-id model to grow and adapt
continuously to progressively available up-to-date labelled data. 
Existing re-id methods can only afford
model re-training from scratch, causing both high
computational cost and response latency to a user. 
They are thus unsuitable for human-in-the-loop model adaptation.

In this work, we solve the two issues by formulating person
re-id as a {\em regression} problem \citep{hoerl1970ridge}. 
Unlike existing methods designed to learn {\em collectively} from all
the training identities a {\em generic} feature embedding space
optimised for classification, verification or ranking, 
we propose to construct an {\em individually} semantic feature embedding space
for identity regression optimised on {\em each training identity}, 
referred to as an {\em Identity Regression
Space} (IRS) defined by all training identity classes. 
Each dimension of IRS corresponds to a specific training person class,
i.e. all training images of the same
identity class are represented by a single unit
vector lying in one unique dimension (axis).
Our modelling objective is therefore to 
train a regression model that maps (embeds) the original image
feature space to this identity regression space. 

We formulate a re-id incremental learning framework
with three fundamental advantages:
{\em First}, it allows quicker re-id system deployment 
after learning from only a small amount of labelled data.
{\em Second}, the learned re-id model facilitates
the subsequent labelling tasks by providing human
a ranking order of unlabelled samples with the labelling targets (i.e. true matches)
in top ranks at high likelihoods.
This reduces manual search time and effort
as compared to the conventional exhaustive eye-balling of unstructured person images.
{\em Third}, the re-id model progressively improves from new labelled data
to further facilitate future labelling. 
This interactive effect is cumulative in a loop: 
More frequently the model updates, more benefit we obtain 
in both reducing labelling effort
and increasing model deployment readiness.

Our {\bf contributions} are three-folds: 
{\bf (1)} We propose the concept of an {\em Identity Regression Space} (IRS)
by formulating re-id as a regression problem for tackling 
the inherent Small Sample Size (SSS) challenge.
This is in contrast to existing methods relying on
classification, verification, or ranking learning spaces 
which are subject to the SSS problem.
The IRS model is featured by an efficient closed-form 
feature embedding solution
without the need for solving an expensive eigen-system and
alternative optimisation.
{\bf (2)} 
We introduce an incremental learning algorithm for efficient on-line IRS model update. 
This facilitates rapidly updating a IRS re-id
model from piecewise new data {\em only}, 
for progressively accommodating update-to-date labelled data
and viewing condition dynamics, 
hence avoiding less efficient model re-training from scratch. 
{\bf (3)} 
We develop an active learning algorithm for
more cost-effective IRS model update with human-in-the-loop,
an under-studied aspect in existing re-id methods. %
Extensive experiments on four popular
datasets VIPeR \citep{VIPeR}, CUHK01 \citep{transferREID},
CUHK03 \citep{Li_DeepReID_2014b} and Market-1501 \citep{zheng2015scalable} 
show the superiority and advantages of the
proposed IRS model over a wide range of state-of-the-art person re-id
models.}



\section{Related Work}
\label{sec:related}
\changed{
\noindent {\bf Person Re-ID. }
Existing person re-id studies focus on two main areas:
feature representation and matching model.
In the literature, a number of hand-crafted image descriptors have 
been designed for achieving general non-learning based view-invariant re-id features
\citep{SDALF_CVPR10,SalienceReId_CVPR13,hanxiao2014GTS,FisherVectorReId_ECCV12,Stan14ColorName,GOG}.
However, these representations alone are often insufficient to accurately 
capture complex appearance variations across cameras.
A common solution is supervised learning 
of a discriminative feature embedding,
subject to classification, pairwise or triplet learning constraints
\citep{Liao_2015_ICCV,wang2014person,wangjoint}.

Our work belongs to the supervised learning based approach
but with a few unique advantages. 
{\em First}, our IRS is designed with each dimension having discriminative semantics, 
rather than learning to optimise. 
We uniquely train a regression mapping from the raw feature space to
the interpretable IRS 
with a close-formed optimisation solution
\citep{hoerl1970ridge,hastie2005elements}
more efficient than
solving eigen-problems 
\citep{liao2015person,zhang2016learning}
and iterative optimisation
\citep{PRDC,Liao_2015_ICCV}.
The IRS addresses the SSS problem
in a similar spirit of 
the NFST re-id model \citep{chen2000new,yu2001direct,zhang2016learning} 
by projecting same-identity images
into a single point.
Importantly, our model uniquely confirms to a well-designed embedding space
rather than relying on intra-person scatter matrix
which may render the solution less discriminative.
%
{\em Second}, 
we further propose an incremental learning algorithm
for sequential model update at new scene and/or dynamic deployments
without model re-training from scratch.
{\em Finally}, 
we investigate active sampling for more cost-effective 
re-id model update. 

\vspace{0.1cm}
\noindent {\bf Subspace Learning. }
	The IRS is a discriminative subspace learning
    method, similar to distance metric learning
	\citep{yang2006distance},
	Fisher Discriminant Analysis (FDA) \citep{fisher1936use,fukunaga2013introduction},
	and cross-modal feature matching
	\citep{Hardoon07cca,sharma2012generalized,kang2015learning}.
	Representative metric learning re-id methods include
	PRDC \citep{PRDC},  
	KISSME \citep{KISSME_CVPR12}, 
	XQDA \citep{liao2015person},
	MLAPG \citep{Liao_2015_ICCV},
	LADF \citep{li2013learning}, and so forth.
	PRDC maximises the likelihood of
	matched pairs with smaller distances than unmatched
	ones. 
	KISSME measures the probability similarity
	of intra-class and inter-class feature differences 
	under the Gaussian distribution assumption.
	sharing the spirit of
	Bayesian Face model \citep{moghaddam2000bayesian}.
%
	KISSME and Bayesian Face are inefficient given high-dimensional features. 
	XQDA overcomes this limitation by uniting
	dimension reduction and metric learning.
	MLAPG tackles the efficiency weakness in learning Mahalanobis function.
	%
	While achieving significant performance gains,
	these methods focus {\em only} on one-time batch-wise model learning
	while ignore incremental learning capability.
	Our model is designed to fill this gap.


\vspace{0.1cm}
\noindent {\bf Incremental Learning. }
Incremental learning (IL) concerns model training from data streams
\citep{poggio2001incremental}. 
%
Often, IL requires extra
immediate on-line model update for making the model ready
to accept new data at any time. 
IL has been explored in 
many different vision tasks, e.g. image classification
\citep{lin2011large,ristin2014incremental}.
%
The closest works w.r.t. our model are 
three re-id methods 
\citep{liu2013pop,wang2016human,martinel2016temporal}.

Specifically, \citet{liu2013pop} consider to
optimise an 
error-prone post-rank search
for refining quickly the ranking lists.
this method is inherently restricted and unscalable due to the need 
for human feedback on all probe images independently.
%
\citet{wang2016human} solves this limitation
by learning  incrementally a unified generalisable re-id model from 
all available human feedback.
\citet{martinel2016temporal} similarly
consider incremental model update in deployment 
for maintaining re-id performance over time. 
Compared to these IL re-id methods, 
the IRS is uniquely characterised with more efficient optimisation
(i.e. a closed-form solution)
with the capability of low response latency. 
%
This is made possible by casting re-id model learning
as a regression problem in the concept of well-design identity embedding space,
in contrast to classification \citep{liu2013pop},
verification \citep{martinel2016temporal},
or ranking \citep{RankSVMReId_BMVC10, wang2016human} learning problem.
Given that all these methods adopt their respective human verification
designs and incremental learning strategies under distinct evaluation settings,
it is impossible to conduct quantitative evaluation among them.

\vspace{0.1cm}
\noindent {\bf Active Learning. }
Active learning (AL) is a strategy for reducing human labelling effort
by selecting most informative samples for annotation 
\citep{settles2012active,kang2004using}.
%
Despite extensive AL studies on generic object classification \citep{osugi2005balancing,cebron2009active,hospedales2012unifying,ebert2012ralf,loy2012stream,kading2015active,wang2016multi},
there exist little re-id attempts
with only two works to our knowledge: 
active person identification \citep{das2015active} and 
temporal re-id adaptation \citep{martinel2016temporal}. 

Specifically,
\citet{das2015active} learn a multi-class
classifier on known identity classes for recognising
training classes, therefore not a re-id model.
Moreover, this model cannot 
support efficient incremental learning as \citep{martinel2016temporal} and IRS,
due to expensive re-training from scratch and hence less suitable
for AL with human in the loop.
\citet{martinel2016temporal}
explore also AL for incremental re-id model update.
In comparison, our AL algorithm is 
more extensive and comprehensive 
(exploitation \& exploration vs. exploitation alone)
with better learning efficiency (no need for iterative optimisation and graph based data clustering).
IRS is thus more suitable for human-in-the-loop driven incremental learning.

\vspace{0.1cm}
\noindent {\bf Ridge Regression. }
Ridge regression \citep{hoerl1970ridge,hastie2005elements}
is one of the most-studied learning algorithms. 
It has an efficient closed-form solution.
with exiting optimised algorithms \citep{paige1982lsqr} readily
applicable to large sized data.  
We ground the IRS re-id model on ridge regression
for inheriting the learning efficiency and scalability advantages, 
Existing attempts for identity verification problems by
class-label regression include~\citep{liao2014open,sharma2012generalized,kang2015learning}. 
\citet{liao2014open} adopted a linear regression based discriminant analysis method
for re-id. 
\citet{sharma2012generalized}
and \citet{kang2015learning} proposed locality regularised 
class-label regression methods for recognition and retrieval.

Beyond these existing works, we systematically explore different
label coding methods, non-linear regression kernelisation, 
model efficiency enhancement and labelling effort minimisation
in an under-studied incremental re-id learning setting.
Moreover, we bridge ridge regression and
FDA \citep{fisher1936use,fukunaga2013introduction}
in feature embedding space design
for more discriminatively encoding
identity sensitive information.  
While the
relationship between FDA and linear regression has been studied
for binary-class \citep{duda2012pattern} and
multi-class \citep{hastie2005elements,park2005relationship}
classification, 
this is the first study that formulates the two jointly in a single framework
for person re-id.

\vspace{0.1cm}
\noindent {\bf Data Scarcity. } 
There are other generic approaches to solving
the SSS challenge.
Two common schemes are domain transfer 
\citep{Transfer_Ryan_ARTEMIS13,Transfer_ICCV13,peng2016unsupervised,geng2016deep,li2017person}
and data augmentation (synthesis)
\citep{mclaughlin2015data,zheng2017unlabeled}.
The former relies on auxiliary data 
(e.g. ImageNet or other re-id datasets)
while the latter generates additional training data
both for enriching the discriminative information
accessible to model training.
Conceptually, they are complementary to the proposed IRS 
with the focus on 
learning a more discriminative embedding space
on the given training data 
from either scratch or pre-trained models.
As shown in our evaluation,
these approaches can be jointly deployed for 
further improving model generalisation (Table \ref{tab:TL}).}

\begin{figure*} [h]
	\centering
	\includegraphics[width=0.99\linewidth]{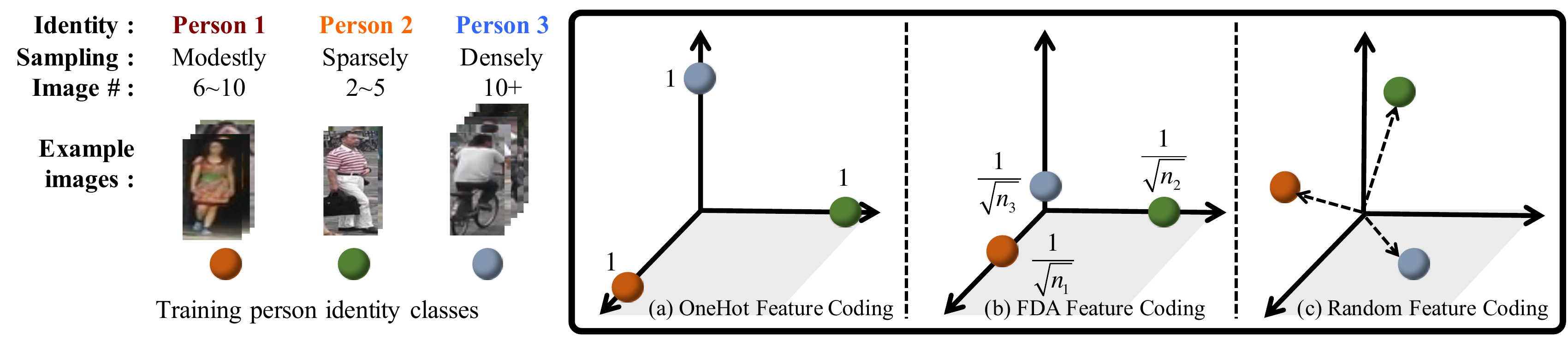}
	\vskip -0.3cm
	\caption{
		Illustration of feature embedding spaces obtained by three training class coding methods. 
		Note, $n_i$ in (b) refers to the training image number of person $i$ extracted from any cameras.
	}
	\label{fig:coding}
\end{figure*}

\begin{figure} 
	\centering
	\includegraphics[width=0.99\textwidth]{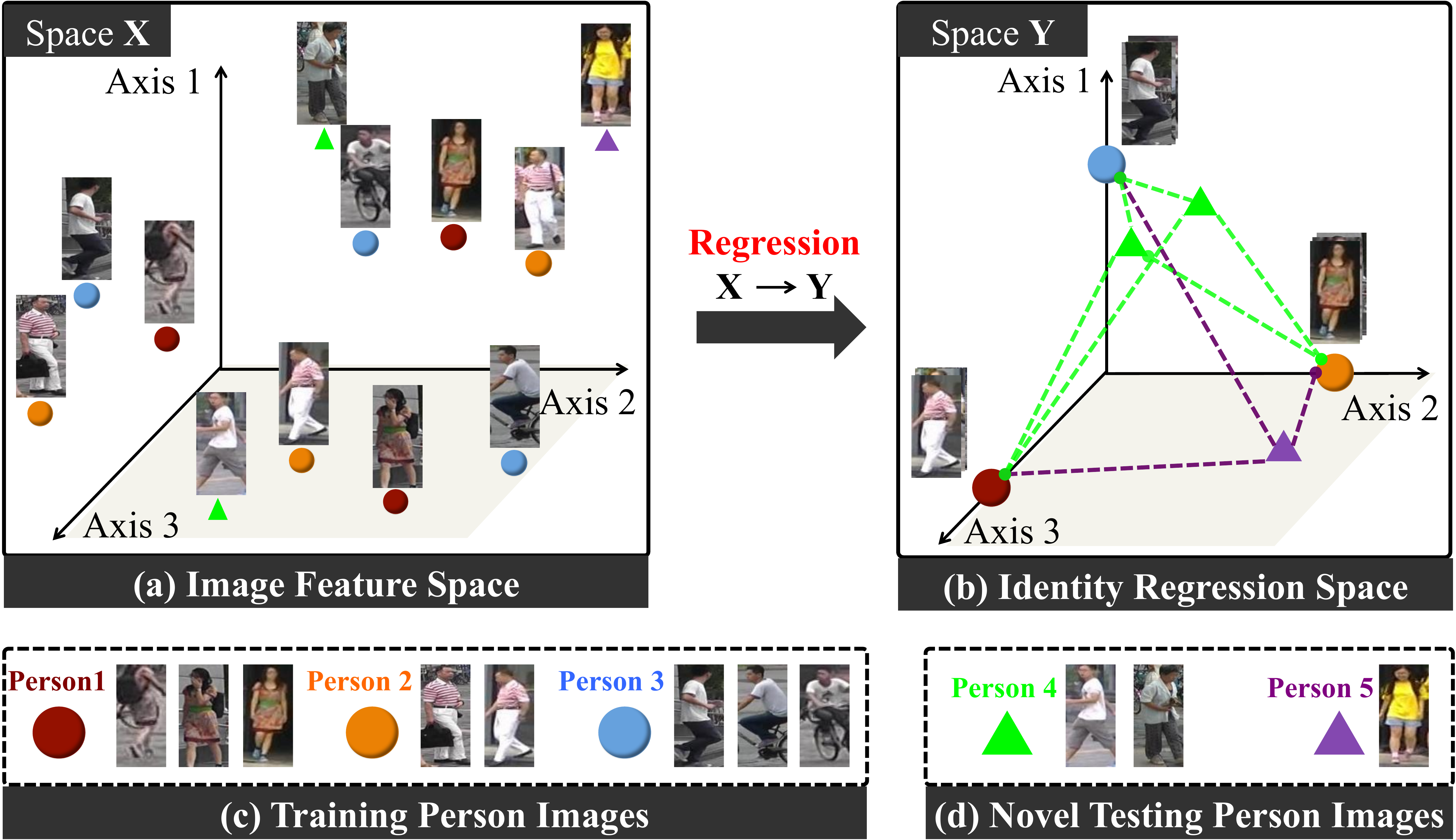}
	\vskip -0.4cm
	\caption{Illustration of our Identity Regression Space (IRS) person re-id model. 
		During model training, by regression we learn an identity discriminative feature embedding from
		(a) the image feature space to (b) the proposed identity regression space
		defined by (c) all training person classes (indicated by circles).
		During deployment, we can exploit the learned feature embedding
		to re-identify (d) novel testing person identities (indicated by triangles)
		in IRS.
	}
	\label{fig:reid_pipeline}
\end{figure}

\section{Identity Regression}
\label{sec:HER}

\subsection{Problem Definition}
\label{sec:problem}
We consider the image-based person re-identification problem \citep{gong2014person}. 
The key is to overcome the unconstrained person appearance variations
caused by significant discrepancy in camera viewing condition and human pose
(Fig. \ref{fig:problem}).
To this end, we aim to 
formulate a feature embedding model for effectively and efficiently
discovering identity discriminative information of cross-view person images. 

Formally, we assume a labelled training dataset
$\bm{X} = [\bm{x}_1,\cdots, \bm{x}_i, \cdots, \bm{x}_n] \in \mathbb{R}^{d\times n}$
where
$\bm{x}_i \in \mathbb{R}^{d \times 1}$ denotes the $d$-dimensional feature vector of image $\bm{x}_i$,
with the corresponding identity label vector
$\bm{l} = [l_1, \cdots, l_i, \cdots, \l_n] \in \mathbb{Z}^{1 \times n}$,
where
$l_i \in \{1,\cdots,c\}$ represents the identity label of image $\bm{x}_i$
among a total of $c$ identities. 
So, these $n$ training images describe $c$ different persons captured under multiple camera views.
We omit the camera label here for brevity.
The model learning objective is to obtain a discriminative feature embedding $\bm{P} \in \mathbb{R}^{d\times m}$, 
i.e. in the embedding space, the distance between intra-person images is 
small whilst that of inter-person images is large
regardless of their source camera views.
In most existing works, the above criterion of compressing intra-person distributions and expanding inter-person distributions is encoded as classification / verification / ranking losses and then a feature embedding is learned by optimising the corresponding objective formulation.
However, due to the SSS problem, the learned embedding space is often
suboptimal and less discriminative. Also, there is often no clear interpretation on
the learned embedding space.

Our method is significantly different: 
Prior to the model training, we first explicitly define an {\em ideal
 feature embedding space}, and then train a regression from the raw feature
space to the defined embedding space.  
The learned regression function is our discriminative feature embedding.
Specifically, we define a set of ``{\em ideal}'' target vectors in the embedding space, denoted by
$\bm{Y} = [\bm{y}_1^\top,\cdots,\bm{y}_n^\top]^\top \in \mathbb{R}^{n \times m}$,
and explicitly assign them to each of the training sample $\bm{x}_i$,
with 
$\bm{y}_i \in \mathbb{R}^{1 \times m}$
referring to $\bm{x}_i$'s target point in the feature embedding space, $i \in \{1,2,\cdots,n\}$
and $m$ referring to the feature embedding space dimension.
In model training, we aim to obtain an optimal feature embedding $\bm{P}$ 
that transforms the image feature $\bm{x}$ 
into its mapping $\bm{y}$ 
with labelled training data $\bm{X}$.
%
During model deployment, 
given a test probe image $\tilde{\bm{x}}^p$ 
and a set of test gallery images $\{\tilde{\bm{x}}_i^g\}$,
we first transform them into the embedding space with the learned feature embedding $\bm{P}$, denoted as
$\tilde{\bm{y}}^p$ and $\{\tilde{\bm{y}}_i^g\}$ respectively. 
Then, we compute the pairwise matching distances between $\tilde{\bm{y}}^p$
and $\{\tilde{\bm{y}}_i^g\}$ by the Euclidean distance metric.
Based on matching distances, we rank all gallery images in ascendant order. 
Ideally, the true match of the probe person is supposed to appear among top ranks.

\subsection{Identity Regression Space}
\label{sec:regression}
To learn an optimal regression function as feature embedding, 
one key question in our framework is how to design the target ``{\em ideal}'' embedding space, in other words, how to set $\bm{Y}$. 
We consider two principles in designing distribution patterns of training samples in the 
embedding space:
\begin{enumerate}
	\item {\em Compactness: } 
	This principle concerns image samples belonging to the {\em same person class}. Even though each person's intra-class distributions may be different in the raw feature space, we argue that in an optimal embedding space for re-id, the variance of all intra-class distributions should be suppressed. Specifically, for every training person, regardless of the corresponding sample size, all samples should be collapsed to a single point 
	so that the embedding space becomes maximally discriminative with respect to person identity. 
	\item {\em Separateness: } 
	This principle concerns image samples belonging to the {\em different person classes}. 
	Intuitively, the points of different person identities should be maximally separated in the embedding space. 
	With a more intuitive geometry explanation, these points should be located on the vertices of a regular simplex  with equal-length edges, 
	so that the embedding space treats equally any training person with a 
	well-separated symmetric structure. 
\end{enumerate}

Formally, 
we assign a unit-length vector on each dimension axis in the feature embedding space
to every training person identity,
i.e. 
we set $\bm{y}_i = [y_{i,1}, \cdots, y_{i,m}]$ 
for the $i$-th training person (Fig. \ref{fig:coding}(a)) as:
\begin{equation}
{y}_{i,j} = 
\begin{cases} 
1,  & \mbox{if } \;\; l_i=j; \\ 
0, & \mbox{if } \;\; l_i\neq j. 
\end{cases}
\quad \mbox{with} \quad
j \in [1,2,\cdots,m],
\label{eq:onehot_Y}
\end{equation}
where $l_i$ is the identity label of image $\bm{x}_i$.
We name this way of setting $\bm{Y}$ as {\em OneHot Feature Coding}.
The embedding space  defined by Eq.~\eqref{eq:onehot_Y}
has a few interesting properties:
\begin{enumerate}
	\item Each dimension in the embedding space corresponds to one specific training person's identity; 
	
	\item  Training persons are evenly distributed in the embedding space and the distances between any two training persons are identical;
	
	\item Geometrically, the points of all training person identities together form a standard simplex.
\end{enumerate}
Because each dimension of this embedding space can be now 
interpreted by one specific training identity, we call such an
embedding space an {\em identity regression space}.
Having the {identity regression space} defined by
Eq.~\eqref{eq:onehot_Y}, we propose to exploit the multivariate ridge
regression algorithm \citep{hoerl1970ridge,zhang2010regularized}. 

In particular, by treating $\bm{Y}$ as the regression output and $\bm{P}$ as the to-be-learned parameter,
we search for a discriminative projection by minimising the mean squared error as:
\begin{equation}
\bm{P}^* = \arg\min_{\bm{P}} \; \frac{1}{2}\|\bm{X}^\top\bm{P} - \bm{Y}\|_F^2 + \lambda\|\bm{P}\|_F^2,
\label{eq:HER_obj}
\end{equation}
where
$\|\cdot\|_F$ is the Frobenius norm,
$\lambda$ controls the regularisation strength.
Critically, this formulation has an efficient closed-form solution:
\begin{equation}
\bm{P}^* = \big(\bm{X}\bm{X}^\top + \lambda\bm{I}\big)^\dagger\bm{X}\bm{Y},
\label{eq:HER_solution}
\end{equation}
where $(\cdot)^\dagger$ denotes the Moore-Penrose inverse, and $\bm{I}$ the identity matrix.
Since our model learning is by regression towards
a training identity space,
we call this method the ``Identity Regression Space'' ({\bf IRS}) model (Fig. \ref{fig:reid_pipeline}).
The IRS re-id feature learning requirement leads naturally to 
exploiting the ridge regression method
for learning the mapping between image features and
this semantic identity space. The novelty of this approach is not in
Eq.~(2) itself, but the IRS learning concept in the re-id context.
Note that,
\changed{we do not select deep models \citep{xiao2016learning} in our IRS implementation
	due to their intrinsic weakness for model incremental learning. 
	Nevertheless, in our experiments we also evaluated
	IRS with a deep learning model (Section 5.1, IV and V).
	Technically, 
	OneHot based IRS {\em feature coding} and {\em embedding} differs fundamentally
	from deep learning classification models due to two modelling differences:
(1) Whilst the latter adopts one-hot {\em class label vectors}, 
the underlying optimised deep features (e.g. the feature layer outputs) 
are not of one-hot style, i.e. not an IRS embedding.
(2) A single softmax prediction may correspond to multiple different logit (i.e. feature) inputs. 
Specifically, even if two logit inputs are different, 
as long as the corresponding element is {\it relatively} larger than others, 
both their softmax outputs will be close to the same one-hot vector. 
In other words, for deep classification models the underlying feature representations of each class are not unique. Therefore, deep classification model are trained under a weaker learning constraint than the IRS
whose feature embedding is trained strictly with only one ground-truth feature vector per class.
The regression algorithm selection is independent of the generic IRS concept.}
	%

\vspace{0.1cm}
\noindent {\bf Remark.}
\sgg{Unlike Fisher Discriminant Analysis \citep{fisher1936use}, 
the proposed IRS has no need for the intra-class and between-class scatter
matrices. 
This renders our model more suitable for addressing the Small Sample Size (SSS) problem
since the intra-class scatter matrix of sparse training data will become
singular, 
which results in computational difficulty \citep{fukunaga2013introduction}.
To solve this SSS problem, one straightforward approach 
is performing dimensionality reduction 
(e.g. principal component analysis) 
before model learning \citep{CVPR13LFDA}.
This however may cause the loss of discriminative power.
An alternative method is directly rectifying the intra-class scatter
by adding a non-singular regularisation matrix 
\citep{mika1999fisher,xiong2014person,liao2015person}. 
Nonetheless, both approaches as above suffer from the degenerate eigenvalue problem 
(i.e. several eigenvectors share the same eigenvalue),
which makes the solution sub-optimal with degraded discrimination \citep{zheng2005foley}.
As a more principled solution, the Null Foley–Sammon Transform (NFST) 
modifies the Fisher discriminative criterion --
Finding null projecting directions 
on which the intra-class distance is zero 
whilst the between-class distance is positive --
so that more discriminant projections corresponding to 
the infinitely large Fisher criterion can be 
obtained \citep{chen2000new,guo2006null}.
The NFST has also been recently employed to solve
the SSS problem in re-id \citep{zhang2016learning}.
While reaching the largest Fisher objective score via exploiting the null space 
of intra-class scatter matrix by NFST,
the between-class scatter is not maximised 
and therefore still an incomplete Fisher discriminative analysis.
It is easy to see that the proposed IRS model shares the spirit of NFST 
in terms of projecting same-class images into a single point
in order to achieve the extreme class {\em compactness} 
and most discriminative feature embedding.
However, unlike the NFST's positive between-class scatter constraint 
-- a weaker optimisation constraint likely resulting in lower discriminative power, 
the model proposed here optimises instead the between-class {\em separateness}
by enforcing the orthogonality between any two different person classes 
in the target feature space to maximise the class 
discrimination and separation in a stronger manner.
%
In terms of model optimisation,
we resort to the more efficient ridge regression paradigm
rather than the Fisher criterion.
Overall, we consider that 
our IRS conceptually extends the NFST by inheriting its local compact classes merit
whilst addressing its global class distribution modelling weakness
in a more efficient optimisation framework.
In our evaluations,
we compare our IRS model with 
the NFST and show the advantages from this new formulation
in terms of both model efficiency and discriminative power.
}

%

\vspace{0.2cm}
\noindent {\bf Alternative Feature Coding.}
Apart from the OneHot feature coding (Eq.~\eqref{eq:onehot_Y}),
other designs of the embedding space can also be readily incorporated into our IRS model. 
We consider two alternative feature coding methods.
The first approach respects the Fisher Discriminant Analysis (FDA)
\citep{fisher1936use,fukunaga2013introduction} criterion, named {\em
FDA Feature Coding}, which is adopted in the preliminary version of this work \citep{wang2016highly}.
Formally, the FDA criterion can be encoded into
our IRS model by
setting target identity regression space as (Fig. \ref{fig:coding}(b)):
%
\begin{equation}
{y}_{ij} = 
\begin{cases} 
\frac{1}{\sqrt{n_i}},  & \mbox{if } \;\; l_i=j; \\ 
0, & \mbox{if } \;\; l_i\neq j. 
\end{cases}
\quad \mbox{with} \quad
j \in [1,2,\cdots,m].
\label{eq:FDA_Y}
\end{equation}
where $n_i$ and $l_i$ refers to the total image number and identity label 
of training person $i$.
A detailed derivation is provided in Appendix \ref{sec:app}.
As opposite to Eq. \eqref{eq:onehot_Y} which treats each person identity equally
(e.g. assigning them with unit-length vectors in the embedding space), 
this FDA coding scheme assigns variable-length vectors
with the length determined by $n_i$. 
As shown in (Fig. \ref{fig:coding}(b)), with the FDA criterion, the resulting training identity simplex in the embedding space is no longer regular.
This may bring benefits for typical classification problems 
by making size-sensitive use of available training data for modelling individual classes
as well as possible, 
but not necessarily for re-id. Particularly,
modelling training classes in such a biased way may instead 
hurt the overall performance since the re-id model is differently required to 
generalise the knowledge from training person classes 
to previously unseen testing ones other than within the training ones 
as in conventional classification.

The second alternative is {\em Random Feature Coding}. That is, we allocate for each training identity a $m$-dimensional random vector
with every element following a uniform distribution over the range of [$0,1$] (Fig. \ref{fig:coding}(c)).
Random coding has shown encouraging effect in 
shape retrieval \citep{zhu2016heat}
and face recognition \citep{zhang2013random}.
In this way, individual dimensions are no longer identity-specific
and training identity regression space are shared largely irregularly.
We will evaluate the effectiveness of these three feature coding methods in Sec. \ref{sec:eval_batch}.
\subsection{Kernelisation}
\label{sec:kernel_IRS}
Given complex variations in viewing condition across cameras, 
the optimal subspace may not be obtainable by linear projections.   
Therefore, we further kernelise the IRS model (Eq. \eqref{eq:HER_solution})
by projecting the data from the original visual feature space 
into a reproducing kernel Hilbert space
$\mathcal{H}$ with an implicit feature mapping function
$\phi(\cdot)$. 
The inner-product of two data points in $\mathcal{H}$
can be computed by a kernel function: 
$h_\text{k}(\bm{x}_i, \bm{x}_j) = \left \langle \phi(\bm{x}_i), \phi(\bm{x}_j) \right \rangle$.
By $h_\text{k}$ (we utilised the typical RBF or Gaussian kernel in our implementation), 
we obtain a kernel representation $\bm{K}\in \mathbb{R}^{n\times n}$,
based on which a corresponding non-linear projection solution can be induced as: 
\begin{equation}
\bm{Q}^* = \big( \bm{K}\bm{K}^\top + \lambda\bm{K} \big)^\dagger\bm{K}\bm{Y}.
\label{eq:ker}
\end{equation}
Once test samples are transformed into 
the kernel space with $h_k$, we can similarly apply the learned projection $\bm{Q}^*$
as the linear case. 
\eddy{We use the kernel version throughout all experiments due to its capability
of modelling the non-linearity which is critical for open space re-id in images
with complex person appearance variations across camera views.}

\begin{figure*} 
	\centering
	\includegraphics[width=0.99\textwidth]{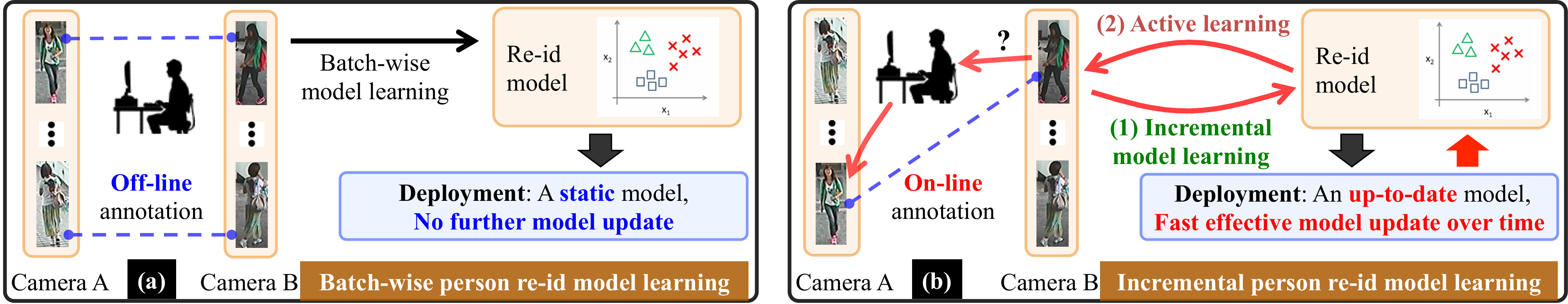}
	\vskip -0.4cm
	\caption{Illustration of different person re-id model learning settings.
		{\bf (a)} Batch-wise person re-id model learning: 
		A re-id model is first learned on an exhaustively labelled training set, 
		and then fixed for deployment without model update;
		{\bf (b)} Incremental person re-id model learning: 
		Training samples are collected
		sequentially on-the-fly with either random or active unlabelled data selection,
		and the re-id model keeps up-to-date by efficient incremental learning
		from the newly labelled data over time.
	}
	\label{fig:batch_incremental}
\end{figure*}

\section{Incremental Identity Regression}
\label{sec:incHER}

In Sec. \ref{sec:HER}, we presented the proposed IRS person re-id model.
Similar to the majority of conventional re-id methods, 
we assume a batch-wise model learning setting:
First collecting all labelled training data and then learning the 
feature embedding model (Fig. \ref{fig:batch_incremental} (a)).
In real-world scenario, however, data annotation is likely to arrive in sequence
rather than at one time particularly when deployed to new arbitrary scenes.
In such case, a practical system requires the incremental learning capability
for cumulatively learning and updating the re-id model over deployment process (Fig. \ref{fig:batch_incremental} (b)-(1)).
On the other hand, incremental learning is essential for temporal model adaptation,
e.g. handling the dynamics in the deployment context \citep{martinel2016temporal}.
A simple and straightforward scheme is to re-train the model from scratch using 
the entire training dataset whenever any newly labelled samples
become available.
Obviously, this is neither computational friendly nor scalable 
particularly for resource/budget restricted deployment.

To overcome this limitation, we introduce an incremental
learning algorithm, named IRS$^\text{inc}$,
for enabling fast model update without the need for re-training from scratch.
%
Suppose at time $t$,
we have
the feature matrix $\bm{X}_t \in \mathbb{R}^{d\times n_t}$ 
of $n_t$ previously labelled images of $c_t$ person identities,
along with $\bm{Y}_t \in \mathbb{R}^{n_t \times m}$ their indicator matrix defined by Eq.~\eqref{eq:onehot_Y}.
We also have the feature matrix $\bm{X}' \in \mathbb{R}^{d\times n'}$  of $n'$ newly labelled
images of $c'$ new person classes,
with $\bm{Y}' \in \mathbb{R}^{n' \times (c_t+c')}$
the corresponding indicator matrix similarly defined by Eq.~\eqref{eq:onehot_Y}.
After merging the new data, 
the updated feature and identity embedding matrix can be represented as:
\begin{equation}
\bm{X}_{t+1} = [\bm{X}_t, \; \bm{X}'],
\quad
\bm{Y}_{t+1} =
\Big[ \begin{array}{c}
\bm{Y}_t\oplus\bm{0}\\
\bm{Y}'  \end{array} \Big],
\label{eq:XY}
\end{equation}
where 
$(\cdot)\oplus\bm{0}$ denotes the matrix augmentation operation, i.e.
padding an appropriate number of zero columns on the right.
%
%
By defining
\begin{equation}
\bm{T}_t = \bm{X}_t\bm{X}_t^\top, 
\end{equation}
and applying Eq. \eqref{eq:XY}, we have 
\begin{equation}
\bm{T}_{t+1} = \bm{T}_t + \bm{X}'\bm{X}'^{\top}.
\label{eq:T}
\end{equation}
For initialisation, i.e. when $t=0$, 
we set $\bm{T}_0 = \bm{X}_0\bm{X}_0^\top + \lambda\bm{I}$.
Also, we can express the projection $\bm{P}_t \in \mathbb{R}^{d \times m}$ 
(Eq.~\eqref{eq:HER_solution}) of our IRS model at time $t$ as 
\begin{equation}
\bm{P}_{t}   = \bm{T}_{t}^\dagger \bm{X}_{t} \bm{Y}_{t}.
\label{eq:P_short}
\end{equation}
%
Our aim is to obtain the feature embedding $\bm{P}_{t+1}$, which
requires to compute $\bm{T}_{t+1}^{\dagger}$.
This can be achieved by applying the Sherman-Morrison-Woodbury formula \citep{woodbury1950inverting} 
to Eq.~\eqref{eq:T} as:
\begin{equation}
\bm{T}_{t+1}^{\dagger} = \bm{T}_t^\dagger - \bm{T}_t^\dagger
\bm{X}'\big(\bm{I}+\bm{X}'^{\top}\bm{T}_t^\dagger\bm{X}'\big)^\dagger
\bm{X}'^{\top}\bm{T}_t^\dagger.
\label{eq:Tdagger}
\end{equation}
Eq.~\eqref{eq:HER_solution} and Eq.~\eqref{eq:XY} together give us:
\begin{align}
\bm{P}_{t+1} &= \bm{T}_{t+1}^\dagger \bm{X}_{t+1} \bm{Y}_{t+1}
\\ \nonumber
&= (\bm{T}_{t+1}^\dagger \bm{X}_t \bm{Y}_t)\oplus\bm{0}
+ \bm{T}_{t+1}^\dagger \bm{X}'\bm{Y}'.
\end{align}
Further with 
Eq.~\eqref{eq:Tdagger} and
Eq.~\eqref{eq:P_short},
we can update $\bm{P}$ as:
\begin{align}
\small
\label{eq:update}
\bm{P}_{t+1}  = \Big(\bm{P}_t - \bm{T}_t^\dagger
\bm{X}'\big(\bm{I}+\bm{X}'^{\top}\bm{T}_t^\dagger\bm{X}'\big)^\dagger
\bm{X}'^{\top}\bm{P}_t\Big) \oplus \bm{0} \\ \nonumber 
+ \bm{T}_{t+1}^\dagger \bm{X}'\bm{Y}'.
\end{align}
Note, the model update (Eq. \eqref{eq:Tdagger} and Eq. \eqref{eq:update}) 
only involves newly coming data samples.
Hence, our method does not require to store the training data
once used for model update.
As only cheap computational cost is involved in such linear operations,
the proposed algorithm well suits for 
on-line responsive re-id model learning
and updating in deployment at large scales in reality.
%

\vspace{0.2cm}
\noindent \textbf{Implementation Consideration. }
The IRS$^\text{inc}$ model
supports incremental learning given either a single new sample ($n' \!\!=\!\! 1$) 
or a small chunk of new samples
($n'\!\! \geqslant\!\! 2$). 
If the data chunk size $n'\ll d$ (where $d$ is
the feature dimension), it is faster to 
perform $n'$ separate updates on each new sample
instead of by a whole chunk. The reason is that,
in such a way
the Moore-Penrose matrix inverse in Eq.~\eqref{eq:Tdagger}
and Eq.~\eqref{eq:update} can be reduced to $n'$ separate
scaler inverse operations, which is much cheaper in numerical computation. 

\subsection{Active Learning for Cost-Effective Incremental Update} 
\label{sec:active}

The incremental learning process described above is {\em passive},
i.e. a human annotator is supposed to label randomly chosen data without considering
the potential value of each selected sample in improving the re-id model.
Therefore, data annotation by this random way is likely to contain redundant information 
with partial labelling effort wasted.
To resolve this problem, we explore the active learning idea \citep{settles2012active}
for obtaining more cost-effective incremental re-id model update (Fig. \ref{fig:batch_incremental} (b)-(2)).


\vspace{0.2cm}
\noindent {\bf Active IRS$^\text{inc}$ Overview. }
In practice, we often have access to a large number of {\it unlabelled} images $\widetilde{\mathcal{P}}$
and $\widetilde{\mathcal{G}}$ captured by disjoint cameras. 
Assume at time step $t \in \{1,\cdots,\tau\}$ with $\tau$ defining the
pre-determined human labelling budget, 
we have the up-to-date IRS$^\text{inc}$ model $m_t$ 
(corresponding to the feature embedding $\bm{P}_t$),
along with $\widetilde{\mathcal{P}_t}$ and $\widetilde{\mathcal{G}_t}$
denoting the remaining unlabelled data.
To maximise labelling profit, 
we propose an {\it active labelling} algorithm for IRS$^\text{inc}$ with the main steps as follows:
\begin{enumerate}
	\item An image $\bm{x}_t^p \in \widetilde{\mathcal{P}_t}$ of a new training identity $l_t$
	is {\em actively} selected by model $m_t$, according to its potential usefulness and importance measured by
	certain active sampling criteria (see details below);
	\item A ranking list of unlabelled images $\widetilde{\mathcal{G}}_t$
	against the selected $\bm{x}_t^p$ is then generated by $m_t$ based matching distances;
	\item For the selected $\bm{x}_t^p$, 
	a human annotator is then asked to manually identify the cross-view true matching image $\bm{x}_t^g \in \widetilde{\mathcal{G}}_t$ in the ranking list, 
	and then generate a new annotation ($\bm{x}_t^p$, $\bm{x}_t^g$);
	\item The IRS$^\text{inc}$ re-id model is updated to $m_{t+1}$ 
	(i.e. $\bm{P}_{t+1}$) from the new data annotation $(\bm{x}_t^p, \bm{x}_t^g)$
	by our incremental learning algorithm (Eq. \eqref{eq:Tdagger} and Eq. \eqref{eq:update}).
\end{enumerate}

\begin{figure*} [t]
	\centering
	\includegraphics[width=0.99\textwidth]{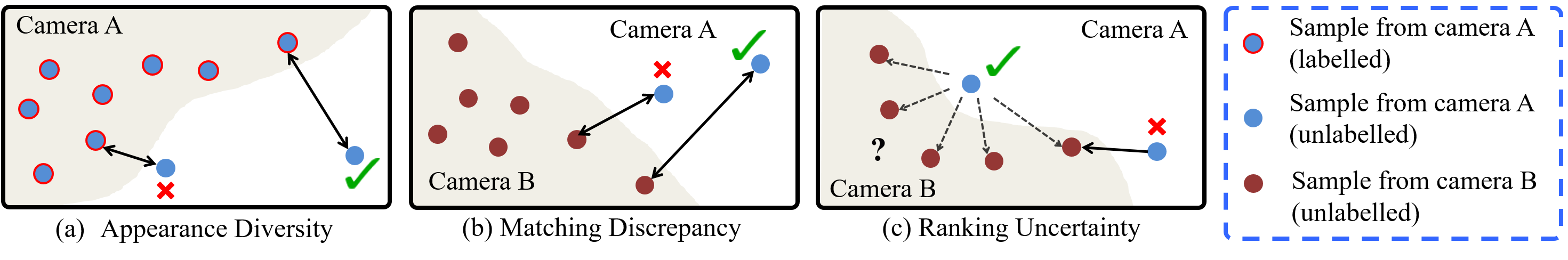}
	\vskip -0.4cm
	\caption{
		Illustration of the proposed active exploration and exploitation selection criteria for more cost-effective incremental re-id model learning.
	}
	\label{fig:active}
\end{figure*}

Among these steps above, the key lies in how to select a good image $\bm{x}_t^p$.
To this end,
we derive a {``Joint Exploration-Exploitation''} ({\bf JointE$^2$})
active sampling algorithm composed of three criteria as follows (Figure~\ref{fig:active}).

\vspace{0.2cm}
\noindent {\bf (I) Appearance Diversity Exploration. }
Intuitively, the appearance diversity of training people
is a critical factor for the generalisation capability of a re-id model.
Thus, the preferred next image to annotate should
lie in the most unexplored region of the population $\widetilde{\mathcal{P}_t}$.
Specifically, at time $t$,
the distance between any two
samples $(\bm{x}_1,\bm{x}_2)$ by the current re-id model is computed as:
\begin{equation}
d(\bm{x}_1,\bm{x}_2|m_t) = (\bm{x}_1 - \bm{x}_2)^\top\bm{P}_t\bm{P}^\top_t(\bm{x}_1 - \bm{x}_2).
\label{eq:dist}
\end{equation}
Given the unlabelled $\widetilde{\mathcal{P}}_t$ and labelled $\mathcal{P}_t$ part of the set $\widetilde{\mathcal{P}}$
($\widetilde{\mathcal{P}}_t \bigcup \mathcal{P}_t = \widetilde{\mathcal{P}}$), 
we can measure the diversity degree of
an unlabelled sample ${\bm{x}}_i^p \in \widetilde{\mathcal{P}}_t$ 
by its distance against the {\it within-view nearest neighbour} in $\mathcal{P}_t$ (Figure~\ref{fig:active} (a)):
\begin{equation}
\begin{aligned} 
\varepsilon_1({\bm{x}}_i^p)
= \min \; d({\bm{x}}_i^p, \bm{x}_j^p|m_t), \\ 
\mbox{s.t.} \;\; {\bm{x}}_i^p \in \widetilde{\mathcal{P}}_t, \;\; \bm{x}_j^p \in \mathcal{P}_t.
\end{aligned}
\label{eq:diversity}
\end{equation}
\sgg{Eq. \eqref{eq:diversity} defines the distance of an
  {\em unlabelled} sample $\bm{x}_i^p$ from the labelled set,
i.e. the distance between $\bm{x}_i^p$ and its nearest labelled sample.
This is not an optimisation operation. 
It is a nearest sample search by ``min'' operation.}
\changed{By maximising the nearest distances}, 
more diverse person appearance can be covered and learned 
for more rapidly increasing the knowledge of the IRS$^\text{inc}$ model, avoiding
repeatedly learning visually similar training samples.

\vspace{0.2cm}
\noindent {\bf (II) Matching Discrepancy Exploration.}
A well learned re-id model is supposed to find the true match of a given image 
with a small cross-view matching distance.
In this perspective, our second criterion particularly
prefers the samples with large matching distances in the embedding space,
i.e. the re-id model $m_t$ remains largely unclear on what are the likely corresponding
cross-view appearances of these ``unfamiliar'' people.
Numerically, we compute the matching distance between an unlabelled sample
${\bm{x}}_i^p \in \widetilde{\mathcal{P}}_t$ and the cross-view true match
(assumed as {\it cross-view nearest neighbour}) in 
$\widetilde{\mathcal{G}}$ (Figure~\ref{fig:active} (b)):
\begin{align}
\varepsilon_2({\bm{x}}_i^p)  & = \min\; d({\bm{x}}_i^p, \bm{x}_j^g|m_t), \\ \nonumber 
& \mbox{s.t.} \;\; {\bm{x}}_i^p \in \widetilde{\mathcal{P}}_t, \;\; \bm{x}_j^g \in \widetilde{\mathcal{G}}.
\end{align}
That is, the unlabelled images with greater $\varepsilon_2({\bm{x}}_i^p)$
are preferred to be selected.


\vspace{0.2cm}
\noindent {\bf (III) Ranking Uncertainty Exploitation. }
Uncertainty-based exploitative sampling schemes have been widely
investigated for classification problems \citep{joshi2009multi,settles2008analysis,ebert2012ralf}.
The essential idea is to query the least certain sample for human to annotate.
Tailored for re-id tasks with this idea, given the similar appearance among different identities,
a weak re-id model may probably generate similar ranking scores for those
visually ambiguous gallery identities with respect to a given probe.
Naturally, it should be useful and informative to manually label such ``challenging'' samples
for enhancing a person re-id model's discrimination power 
particularly with regard to such person appearance (Figure~\ref{fig:active} (c)).
To obtain such person images, 
we define a matching distance based probability distribution 
over all samples
$\bm{x}^g_j \in \widetilde{\mathcal{G}}$ for a given cross-view image 
$\bm{x}^p_i \in \widetilde{\mathcal{P}}$:
\begin{equation}
p_{m_t}(\bm{x}^g_j|\bm{x}^p_i) = \frac{1}{Z_{i}^t}{e^{-d({\bm{x}}_i^p, \bm{x}_j^g|m_t)}}, 
\end{equation}
where
\begin{equation}
Z_{i}^t = {\sum_k e^{-d({\bm{x}}_i^p, \bm{x}_k^g|m_t)}}, \;\; \bm{x}_k^g \in \widetilde{\mathcal{G}}. \nonumber
\end{equation}
The quantity $p_{m_t}(\bm{x}^g_j|\bm{x}^p_i)$ gives a high entropy when most ranking scores are adjacent to each other, indicating great information to mine
from the perspective of information theory \citep{akaike1998information}.
In other words, the model has only a low confidence on its
generated ranking list considering that 
only a very few number of cross-camera samples are likely to be true matches
rather than many of them. 
Consequently, our third criterion is designed as:
\begin{align}
\varepsilon_3({\bm{x}}_i^p) & = - \sum_j p_{m_t}(\bm{x}^g_j|\bm{x}^p_i) \log p_{m_t}(\bm{x}^g_j|\bm{x}^p_i), \\ \nonumber
& \quad \mbox{s.t.} \;\; {\bm{x}}_i^p \in \widetilde{\mathcal{P}}_t, \;\; \bm{x}_j^g \in \widetilde{\mathcal{G}}.
\end{align}
which aims to select out those associated with high model ranking ambiguity.

\begin{algorithm}[t]
	\SetInd{0.1cm}{0.3cm}
	\caption{\small {Active IRS$^\text{inc}$} 
	}
	\KwData{\\
		\quad (1) Unlabelled image set $\widetilde{\mathcal{P}}$ and $\widetilde{\mathcal{G}}$ from disjoint cameras; \\ 
		\quad (2) Regularisation strength $\lambda$; \\
		\quad (3) Labelling budget $\tau$.
	}
	\vspace{0.1cm}
	\KwResult{\\
		\quad (1) Discriminative feature embedding matrix $\bm{P}$;
	}
	\vspace{0.1cm}
	{\bf Initialisation}: \\
	\quad (1) Randomly label a small seed set $\bm{X}_0$, $\bm{Y}_0$; \\
	\quad (2) Set $\bm{T}_0^\dagger=(\bm{X}_0\bm{X}_0^\top + \lambda\bm{I})^\dagger$; \\
	\quad (3) Set $\bm{P}_0 = \bm{T}_0^\dagger\bm{X}_0\bm{Y}_0$ (Eq. \eqref{eq:HER_solution}).\\
	\vspace{0.1cm}
	{\bf Active Labelling}: \\
	\For{$ t = 0:\tau-1$}
	{ 
		(1) Select an unlabelled sample $\bm{x}_t^p \in \widetilde{\mathcal{P}}_t$ (Eq. \eqref{eqn:final_active}); \\
		(2) Rank the images in $\widetilde{\mathcal{G}}_t$ against the selection $\bm{x}_t^p$; \\
		(3) Human annotator verifies the true match in $\widetilde{\mathcal{G}}_t$; \\
		(4) Generate a new annotation $(\mathcal{I}_t^p,\mathcal{I}_t^g)$; \\
		(5) Update $\bm{T}_{t+1}^\dagger$ (Eq. \eqref{eq:Tdagger});\\ 
		(6) Update $\bm{P}_{t+1}$ (Eq. \eqref{eq:update}).
	}
	\Return $\bm{P} = \bm{P}_\tau$;
	\label{algo_AL}
\end{algorithm}

\begin{figure*} [ht!]
	\centering
	\subfigure[VIPeR\citep{VIPeR}]{
		\includegraphics[height=0.25\linewidth,width=0.24\linewidth]{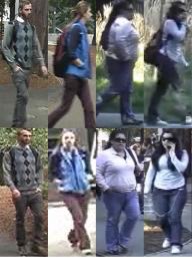}}
	\subfigure[CUHK01\citep{transferREID}]{
		\includegraphics[height=0.25\linewidth,width=0.24\linewidth]{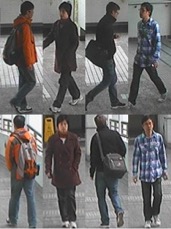}}
	\subfigure[CUHK03\citep{Li_DeepReID_2014b}]{
		\includegraphics[height=0.25\linewidth,width=0.24\linewidth]{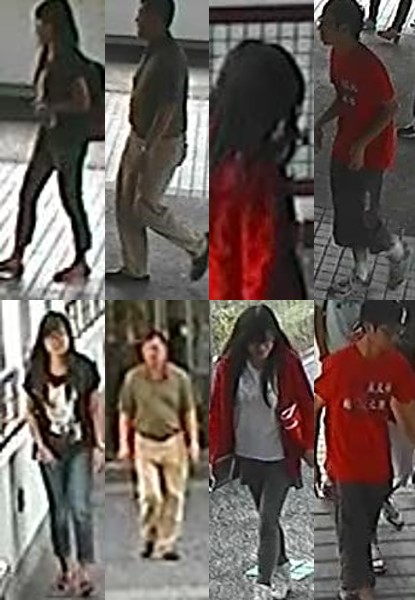}}
	\subfigure[Market-1501\citep{zheng2015scalable}]{
		\includegraphics[height=0.25\linewidth,width=0.24\linewidth]{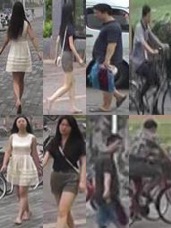}}
	\caption{
		Example person images from four person re-id datasets.
		Two images of each individual columns present the same person.
	}
	\label{fig:dataset}
	\vspace{-0.3cm}
\end{figure*}

\vspace{0.2cm}
\noindent {\bf Joint Exploration-Exploitation. }
Similar to the model in \citep{cebron2009active,ebert2012ralf},
we combine both exploitation and exploration based criteria into our final active selection 
standard, formally as:
\begin{equation}
\label{eqn:final_active}
\varepsilon(\bm{x}^p_i) = \varepsilon_1(\bm{x}^p_i) +
\varepsilon_2(\bm{x}^p_i) + \varepsilon_3(\bm{x}^p_i).
\end{equation}
\eddy{To eliminate scale discrepancy, 
we normalise $\varepsilon_1$, $\varepsilon_2$, $\varepsilon_3$ to the unit range $[0,1]$ respectively before fusing them. 
Specifically, given $\varepsilon_1$ scores of all unlabelled samples, 
we normalise them by dividing the maximal value so that the highest $\varepsilon_1$
is $1$. The same operation is performed on $\varepsilon_2$ and $\varepsilon_3$.}
%
%
%
%

In summary, with Eq. \eqref{eqn:final_active}, all the unlabelled samples in $\widetilde{\mathcal{P}}$
can be sorted accordingly,
and the one with highest $\varepsilon(\bm{x}^p_i)$ is then
selected for human annotation.
An overview of our proposed active learning based incremental model learning and updating 
is presented in Algorithm \ref{algo_AL}.
We will show the effect of our proposed active labelling method
in our evaluations
(Sec. \ref{sec:eval_incre}).

\eddy{
\subsection{Kernelisation}
We kernelise similarly 
the incremental IRS algorithm as in 
Sec. \ref{sec:kernel_IRS}.
Specifically, we first obtain the kernel representation of new training data 
and then conduct model incremental learning in the Hilbert space.
We utilise the kernelised model with
its non-linear modelling power in all 
incremental re-id model learning experiments including 
active sampling with human-in-the-loop.}

\section{Experiments}
\label{sec:experiment}

\noindent \textbf{Datasets.}
For model evaluation, four person re-id benchmarks were used:
VIPeR \citep{VIPeR}, 
CUHK01 \citep{transferREID},  
CUHK03 \citep{Li_DeepReID_2014b},
and Market-1501 \citep{zheng2015scalable}, as summarised in Table \ref{tab:dataset_stats}.
We show in Fig. \ref{fig:dataset} some
examples of person images from these datasets. Note that the datasets were collected with different data
sampling protocols: 
(a) VIPeR has one image per person per view; 
(b) CUHK01 contains two images per person per view; 
(c) CUHK03 consists of a
maximum of five images per person per view, and also provides both manually
labelled and auto-detected image bounding boxes with the latter posing
more challenging re-id test due to unknown misalignment of the
detected bounding boxes; 
(d) Market-1501 has variable numbers of
images per person per view. 
%
%
%
These four datasets present a good selection of re-id test scenarios
with different population sizes under realistic viewing conditions
exposed to large variations in human pose and strong similarities
among different people.

\begin{table} 
	\centering
	\renewcommand{\arraystretch}{0.8}
	\setlength{\tabcolsep}{0.06cm}
	\caption{
		Statistics of person re-id datasets. BBox: Bounding Box.
	}
	\vskip -10pt 
	\begin{tabular}{|c||c|c|c|c|}
		\whline
		Dataset  & 
		{Cameras} & 
		{Persons} & 
		{Labelled BBox} & 
		{Detected BBox} \\ \hline 
		
		VIPeR
		& 2 & 632 & 1,264 & 0 \\  
		CUHK01
		& 2 & 971 & 1,942 & 0 \\
		CUHK03
		& 6 & 1,467 & 14,097 & 14,097 \\
		Market-1501
		& 6 & 1,501 & 0 & 32,668 \\
		\whline
	\end{tabular}
	\label{tab:dataset_stats}
\end{table}

\vspace{0.1cm}
\noindent \textbf{Features. }
To capture the detailed information of
person appearance, we adopted three state-of-the-art feature
representations with variable dimensionalities from 10$^4$ to 10$^2$: 
{\bf (1)} {\it Local Maximal Occurrence} (LOMO) feature \citep{liao2015person}: 
The LOMO feature is based on a HSV colour histogram and Scale
Invariant Local Ternary Pattern \citep{liao2010modeling}. 
For alleviating the negative effects caused by camera view discrepancy,
the Retinex algorithm \citep{land1971lightness} is applied to pre-process person images. 
The feature dimension of LOMO is rather high at $26,960$, therefore
expensive to compute. \\ 
{\bf (2)} {\it Weighted Histograms of Overlapping Stripes} (WHOS) feature \citep{KCCAReid,lisanti2015person}:
The WHOS feature contains HS/RGB histograms and HOG \citep{HOG_PersonDetect_CVPR12} of image grids, with a centre support kernel as weighting to 
approximately segmented person foreground from background clutters. 
We implemented this feature model as described by \citet{KCCAReid}.
The feature dimension of WHOS is moderate at $5,138$. \\
{\bf (3)} {\it Convolutional Neural Network} (CNN) feature \citep{xiao2016learning}: 
Unlike hand-crafted LOMO and WHOS features, 
deep CNN person features are learned from image data. 
Specifically, we adopted the DGD CNN \citep{xiao2016learning} 
and used the FC$_7$ output as re-id features.
The DGD feature has a rather low dimension of $256$, 
thus efficient to extract.
\changed{Following \citet{xiao2016learning},
we trained the DGD by 
combining labelled and detected person bounding box images
(a total $26,246$ images)
with the original authors released codes.}
We then deployed the trained DGD to extract deep features of
the test image data for CUHK03 (the same domain).  
On Market-1501, the CUHK03
trained DGD was further fine-tuned on the $12,936$ Market-1501 training
images for domain adaptation. 
On VIPeR and CUHK01, the CUHK03 trained DGD was {\em directly}
deployed {\em without} any fine-tuning as there are insufficient
training images to make effective model adaptation, with
only $632$ and $1,940$ training images for VIPeR and CUHK01
respectively. 

\vspace{0.1cm}
\noindent {\bf Model Training Settings.}
In evaluations, we considered extensively comparative experiments under 
two person re-id model training settings:
%
{\bf (I)} {\it Batch-wise model training}: 
In this setting, we followed the conventional supervised re-id scheme 
commonly utilised in most existing methods, that is,
first collecting all training data and then learning a re-id model
{\em before} deployment. 
%
{\bf (II)} {\it Incremental model training}: 
In contrast to the batch-wise learning, 
we further evaluated a more realistic data labelling scenario where
more training labels are further collected over time {\em after} model deployment.
The proposed IRS$^{\text{inc}}$ model was deployed for this incremental learning setting.

\subsection{Batch-Wise Person Re-Id Evaluation}
\label{sec:eval_batch}

\noindent {\bf Batch-Wise Re-Id Evaluation Protocol. } 
To facilitate quantitative comparisons with existing re-id methods,
we adopted the standard supervised re-id setting to evaluate the proposed IRS model.
Specifically,
on {\em VIPeR}, we split randomly the whole population of the dataset
(632 people) into two halves:
One for training (316) and another for testing (316).
We repeated $10$ trials of random people splits and utilised the averaged results. 
On {\em CUHK01}, we considered two benchmarking training/test people split settings: 
(1) 485/486 split: randomly selecting $485$ identities for training and
the other $486$ for testing \citep{liao2015person,zhang2016learning}; 
(2) 871/100 split: randomly selecting $871$ 
identities for training and the other $100$ for testing \citep{Ahmed2015CVPR,shi2016embedding}.
As CUHK01 is a multi-shot (e.g. multiple images per person per camera view) dataset, 
we computed the final matching distance between two people 
by averaging corresponding cross-view image pairs.
Again, we reported the results averaged over $10$ random trials for either people split.  
On {\em CUHK03}, following \citet{Li_DeepReID_2014b} we repeated 20 times of random 1260/100 people splits for model training/test and reported the averaged accuracies 
under the single-shot evaluation setting\citep{zhang2016learning}. 
%
On {\em Market-1501}, we used the standard training/test (750/751) people split provided by \citet{zheng2015scalable}. 
On all datasets, we exploited the cumulative matching characteristic (CMC) 
to measure the re-id accuracy performance.
On Market-1501, we also considered the recall measure 
of multiple truth matches
by mean Average Precision (mAP), 
i.e. first computing the area under the Precision-Recall curve for
each probe, then calculating the mean of Average Precision over all
probes \citep{zheng2015scalable}.


In the followings, we evaluated: 
(i) Comparisons to state-of-the-arts, 
(ii) Effects of embedding space design,
(iii) Effects of features,
(iv)  Deep learning regression,
(v) Complementary of transfer learning and IRS,
(vi) Comparisons to subspace/metric learning models,
(vii) Regularisation sensitivity, and
(viii) Model complexity. 

\vspace{0.1cm}
\noindent \textbf{(I) Comparisons to the State-of-The-Arts. }
We first evaluated the proposed IRS model by extensive comparisons 
to the existing state-of-the-art re-id models under the standard
supervised person re-id setting. We considered a wide range of
existing re-id methods, including both hand-crafted and deep learning
models. In the following experiments, we deployed the {\em OneHot Feature
  Coding} (Eq. \eqref{eq:onehot_Y} in Sec. \ref{sec:regression}) for the
identity regression space embedding of our IRS model unless stated otherwise. We
considered both single- and multi-feature based person re-id
performance, and also compared re-id performances of different models
on auto-detected person boxes when available in CUHK03 and Market-1501.


\begin{table} 
	\centering
	\renewcommand{\arraystretch}{0.89}
	\setlength{\tabcolsep}{0.2cm}
	\caption{
		Re-Id performance comparison on the VIPeR benchmark.
		($^*$): Multiple features fusion.
	}
	\vskip -10pt
	\begin{tabular}{|c|cccc|}
		\whline
		Dataset  & \multicolumn{4}{c|}{VIPeR} \\ \hline 
		Rank (\%)     & R1   & R5   & R10   & R20    \\ \hline \hline
		LADF \citep{li2013learning} & 29.3 & 61.0 & 76.0 & 88.1 \\
		MFA \citep{MFA} & 32.2 & 66.0 & 79.7 & 90.6 \\
		kLFDA~\citep{xiong2014person}& 38.6 & 69.2 & 80.4  & 89.2 \\ 
		XQDA~\citep{liao2015person}& 40.0 & 68.1 & 80.5  & 91.1 \\ 
		MLAPG~\citep{Liao_2015_ICCV}& 40.7 & 69.9 & 82.3  & 92.4 \\ 
		NFST~\citep{zhang2016learning}& 42.3 & 71.5 & 82.9  & 92.1\\ 
		%
		%
		LSSCDL \citep{LSSCDL} & 42.7 & - & 84.3 & 91.9 \\ 
		TMA~\citep{martinel2016temporal} & 43.8 & - & 83.8 & 91.5 \\
		HER~\citep{wang2016highly} & 45.1 & 74.6 & 85.1 & 93.3 \\
		\hline 
		DML \citep{yi2014deep} & 28.2 & 59.3 & 73.5 & 86.4 \\
		DCNN+~\citep{Ahmed2015CVPR}& 34.8 & 63.6 & 75.6  & 84.5 \\ 
		SICI~\citep{wangjoint}  & 35.8 & - & -  & -\\ 
		DGD~\citep{xiao2016learning} & 38.6 & - & -  & \\ 
		Gated S-CNN~\citep{Gated_SCNN} & 37.8 & 66.9 &77.4   & - \\
		MCP~\citep{Cheng_TCP} &\bf 47.8 & 74.7 & 84.8 & 91.1\\
		\hline
		\bf IRS~(WHOS) & 44.5 &  \bf 75.0 & \bf 86.3  & \bf 93.6  \\  %
		\bf IRS~(LOMO) & 45.1 & 74.6 &  85.1 & 93.3 \\ 
		\bf IRS~(CNN) & 33.1 &  59.9 &71.5  & 82.2  \\  %
		\hline 
		\hline
		MLF$^*$ \citep{Zhao_MidLevel_2014a} 
		& 43.4 & 73.0 & 84.9 & 93.7 \\
		ME$^*$~\citep{Anton_2015_CoRR} & 45.9 & 77.5 & 88.9  & 95.8  \\
		CVDCA$^*$ \citep{chen2015CVDCA}
		& 47.8 & 76.3 & 86.3 & 94.0 \\
		FFN-Net$^*$ \citep{wu2016enhanced}
		& 51.1 & 81.0 & 91.4 & {\bf 96.9} \\
		NFST$^*$~\citep{zhang2016learning} & 51.2 & 82.1 & 90.5 & 95.9 \\ 
		HER$^*$~\citep{wang2016highly} & 53.0 & 79.8 & 89.6 & 95.5 \\ 
		GOG$^*$ \citep{GOG} & 49.7 & - & 88.7 & 94.5 \\
		SCSP$^*$ \citep{chen2016similarity} &  53.5 & {\bf 82.6} & {\bf 91.5} & 96.7 \\
		\hline
		\bf IRS~(WHOS+LOMO+CNN)$^*$ & \bf 54.6 & 81.5 & 90.3 & 95.7 \\ \whline
	\end{tabular}
	\label{tab:CMC_conventional_VIPeR}
\end{table}

\vspace{0.1cm}
\noindent {\bf\em Evaluation on VIPeR. } 
Table~\ref{tab:CMC_conventional_VIPeR} shows a comprehensive
comparison on re-id performance between our IRS model (and its
variations) and existing models using the VIPeR benchmark
\citep{VIPeR}.
It is evident that our IRS model with a non-deep feature LOMO,
IRS(LOMO), is better than all existing methods\footnote{
	The HER model presented in our preliminary work
        \citep{wang2016highly} is the same as
	IRS(LOMO) with FDA coding (Eq. \eqref{eq:FDA_Y}), 
	i.e. HER = IRS-FDA(LOMO).
	On the other hand, IRS(LOMO) in Tables
        \ref{tab:CMC_conventional_VIPeR},
	\ref{tab:CMC_conventional_CUHK01},
	\ref{tab:CMC_conventional_CUHK03} and
	\ref{tab:CMC_conventional_market} is IRS-OneHot(LOMO).
	The effects of choosing different coding is evaluated later
	 (Table \ref{tab:CMC_TISP}).
} except the deep model
MCP~\citep{Cheng_TCP}, with Rank-1 45.1$\%$ vs. 47.8$\%$ respectively.
Interestingly, using our CUHK03
trained CNN deep feature {\em without} fine-tuning on VIPeR, i.e. IRS(CNN),
does not offer extra advantage (Rank-1 33.1$\%$), due to the significant
domain drift between VIPeR and CUHK03. This becomes more clear when
compared with the CUHK01 tests below. Moreover, 
given a score-level fusion on the matching of three different features,
IRS(WHOS+LOMO+CNN), the IRS can benefit from further boosting on
its re-id performance, obtaining the best Rank-1 rate at 54.6$\%$. 
These results demonstrate the effectiveness of the proposed IRS model in
learning identity discriminative feature embedding because of our {\em
  unique} approach on identity regression to learning a re-id feature embedding
space, in contrast to existing established ideas on classification,
verification or ranking based supervised learning of a re-id model.

\begin{table} 
	\centering
	\renewcommand{\arraystretch}{0.8}
	\setlength{\tabcolsep}{0.17cm}
	\caption{
		Re-id performance comparison on the CUHK01 benchmark.
		($^*$): Multiple features fusion.
	}
	\vskip -10pt
	\begin{tabular}{|c|cccc|}
		\whline
		Dataset  & \multicolumn{4}{c|}{CUHK01 (486/485 split)} \\ \hline 
		Rank (\%)     & R1   & R5   & R10   & R20    \\ \hline \hline
		kLFDA~\citep{xiong2014person}& 54.6 & 80.5 & 86.9  & 92.0 \\ 
		XQDA~\citep{liao2015person}& 63.2 & 83.9 & 90.0  & 94.2 \\ 
		MLAPG~\citep{Liao_2015_ICCV}& 64.2 & 85.4 & 90.8  & 94.9 \\ 
		NFST~\citep{zhang2016learning}& 65.0 & 85.0 & 89.9  & 94.4\\ 
		HER~\citep{wang2016highly}& 68.3 & 86.7 & 92.6  & 96.2\\ 
		\hline
		DCNN+~\citep{Ahmed2015CVPR}& 47.5 & 71.6 & 80.3  & 87.5 \\ 
		MCP~\citep{Cheng_TCP} & 53.7 & 84.3 & 91.0 & 93.3\\
		DGD~\citep{xiao2016learning} & 66.6 & - & - & - \\ 
		\hline	
		\bf IRS~(WHOS) & 48.8 & 73.4 & 81.1 & 88.3 \\
		\bf IRS~(LOMO) & 68.3 & 86.7 &92.6  & 96.2 \\ 
		\bf IRS~(CNN) & \bf 68.6 & \bf 89.3 & \bf 93.9 & \bf 97.2  \\  
		\hline
		\hline
		ME$^*$~\citep{Anton_2015_CoRR} & 53.4 & 76.4 & 84.4  & 90.5  \\
		FFN-Net$^*$ \citep{wu2016enhanced} & 55.5 & 78.4 & 83.7 & 92.6 \\ 
		GOG$^*$ \citep{GOG} & 67.3 & 86.9 & 91.8 & 95.9 \\
		NFST$^*$~\citep{zhang2016learning} & 69.1 & 86.9 & 91.8 & 95.4 \\
		HER$^*$~\citep{wang2016highly} & 71.2 & 90.0 & 94.4 & 97.3 \\
		\hline
		\bf IRS~(WHOS+LOMO+CNN)$^*$ & \bf 80.8 & \bf 94.6 & \bf 96.9 & \bf 98.7 \\ 
		\hline
		\hline
		Dataset  & \multicolumn{4}{c|}{CUHK01 (871/100 split)}  \\ \hline
		FPNN \citep{Li_DeepReID_2014b} & 27.9 & 59.6 & 73.5 & 87.3 \\
		DCNN+~\citep{Ahmed2015CVPR} & 65.0 & - & - & -\\
		JRL \citep{chen2016deep} & 70.9 & 92.3 & 96.9 & 98.7 \\ 
		EDM~\citep{shi2016embedding} & 69.4 & - & -  &- \\ 
		SICI~\citep{wangjoint}  & 71.8 & - & -  & -\\ 
		\hline
		\bf IRS~(WHOS) & 77.0 & 92.8 & 96.5 & 99.2 \\
		\bf IRS~(LOMO) & 80.3 & 94.2 & 96.9  & 99.5 \\ 
		\bf IRS~(CNN) & 84.4 & 98.2 &  \bf 99.8 & \bf 100  \\  
		\bf IRS~(WHOS+LOMO+CNN)$^*$ & \bf 88.4 & \bf 98.8  & 99.6 & \bf 100 \\ \whline
	\end{tabular}
	\label{tab:CMC_conventional_CUHK01}
\end{table}

\begin{table} 
	\centering
	\renewcommand{\arraystretch}{0.85}
	\setlength{\tabcolsep}{0.16cm}
	\caption{
		Re-id performance comparison on the CUHK03 benchmark. 
		($^*$): Multiple features fusion.
	}
	\vskip -10pt
	\begin{tabular}{|c|cccc|}
		\whline
		Dataset  & \multicolumn{4}{c|}{CUHK03 (Manually)} \\ \hline 
		Rank (\%) & R1 & R5 & R10 & R20 \\ \hline \hline
		kLFDA~\citep{xiong2014person}& 45.8 & 77.1 & 86.8  & 93.1 \\ 
		XQDA~\citep{liao2015person}& 52.2 & 82.2 & 92.1  & 96.3 \\ 
		MLAPG~\citep{Liao_2015_ICCV}& 58.0 & 87.1 & 94.7  & 98.0 \\ 
		NFST~\citep{zhang2016learning}& 58.9 & 85.6 & 92.5  & 96.3\\ 
		HER~\citep{wang2016highly}& 60.8 & 87.0 & 95.2  & 97.7 \\
		\hline
		DCNN+~\citep{Ahmed2015CVPR} & 54.7 & 86.5 & 93.9  & \bf 98.1 \\ 
		EDM~\citep{shi2016embedding} & 61.3 & - & -  &- \\
		DGD~\citep{xiao2016learning} & 75.3 & - & -  & \\ 
		\hline
		\bf IRS~(WHOS) & 59.6 & 87.2 & 92.8 & 96.9 \\
		\bf IRS~(LOMO) & 61.6 & 87.0 & 94.6  & 98.0 \\ 
		\bf IRS~(CNN) &\bf 81.5 &\bf 95.7 &\bf 97.1  & 98.0  \\  %
		\hline
		\hline
		ME$^*$~\citep{Anton_2015_CoRR} & 62.1 & 89.1 & 94.3  & 97.8  \\ 
		NFST$^*$~\citep{zhang2016learning} & 62.6 & 90.1 & 94.8 & 98.1 \\
		HER$^*$~\citep{wang2016highly}& 65.2 & 92.2 & 96.8  & \bf 99.1 \\
		GOG$^*$ \citep{GOG} & 67.3 & 91.0 &  96.0 & - \\
		\hline
		\bf IRS~(WHOS+LOMO+CNN)$^*$ & \bf 81.9 & \bf 96.5 & \bf 98.2 & 98.9
		\\ \hline \hline
		Dataset  & \multicolumn{4}{c|}{CUHK03 (Detected)} \\ \hline
		%
		KISSME \citep{KISSME_CVPR12} & 11.7 & 33.3 & 48.0 & - \\
		XQDA~\citep{liao2015person}& 46.3 & 78.9 & 83.5  & 93.2 \\ 
		MLAPG~\citep{Liao_2015_ICCV}& 51.2 & 83.6 & 92.1  & 96.9 \\ 
		L$_1$-Lap~\citep{ElyorECCV16} & 30.4 & - & - & - \\
		NFST~\citep{zhang2016learning}& 53.7 & 83.1 & 93.0  & 94.8\\ 
		\hline
		DCNN+~\citep{Ahmed2015CVPR}& 44.9 & 76.0 & 83.5  & 93.2 \\ 
		EDM~\citep{shi2016embedding} & 52.0 & - & -  &- \\ 
		SICI~\citep{wangjoint}  & 52.1 & 84.9 & 92.4  & -\\ 
		S-LSTM~\citep{S_LSTM} & 57.3 & 80.1 & 88.3 & -\\ 
		Gated S-CNN~\citep{Gated_SCNN} & 68.1 & 88.1 & 94.6  & - \\ 
		\hline
		\bf IRS~(WHOS) & 50.6 & 82.1 & 90.4 & 96.1 \\
		\bf IRS~(LOMO) & 53.4 & 83.1 &   91.2 & 96.4 \\ 
		\bf IRS~(CNN) & \bf 80.3 & \bf 96.3 & \bf 98.6  & \bf 99.0  \\  %
		\hline
		\hline
		NFST$^*$~\citep{zhang2016learning} & 54.7 & 84.8 & 94.8 & 95.2 \\
		GOG$^*$ \citep{GOG} & 65.5 & 88.4 & 93.7 & - \\
		\hline
		\bf IRS~(WHOS+LOMO+CNN)$^*$ & \bf 83.3  & \bf 96.2 & \bf 97.9 & \bf 98.6 \\ \whline
	\end{tabular}
	\label{tab:CMC_conventional_CUHK03}
\end{table}

\begin{table} 
	\centering
	\renewcommand{\arraystretch}{0.9}
	\setlength{\tabcolsep}{0.13cm}
	\caption{
		Re-id performance comparison on the Market-1501 benchmark.
		($^*$): Multiple features fusion.
	}
	\vskip -10pt
	\begin{tabular}{|c|cc|cc|}
		\whline
		Dataset  & \multicolumn{4}{c|}{Market-1501} \\ \hline 
		Query Per Person   & \multicolumn{2}{c|}{Single-Query}  & \multicolumn{2}{c|}{Multi-Query}    
		\\ \hline
		Metric (\%)   & R1   & mAP   & R1   & mAP    \\ \hline \hline
		KISSME \citep{KISSME_CVPR12} & 40.5 & 19.0 & - & - \\
		MFA \citep{MFA} & 45.7 & 18.2 & - & - \\
		kLFDA \citep{xiong2014person} & 51.4 & 24.4 & 52.7 & 27.4 \\
		XQDA \citep{liao2015person}  &43.8 & 22.2 & 54.1 & 28.4\\
		SCSP~\citep{chen2016similarity} & 51.9 & 26.3 & - & - \\
		NFST~\citep{zhang2016learning} & 55.4 & 29.9 & 68.0 & 41.9\\
		TMA~\citep{martinel2016temporal} & 47.9 & 22.3 & - & - \\
		\hline
		SSDAL~\citep{su2016deep} & 39.4 & 19.6 & 49.0 & 25.8 \\
		S-LSTM~\citep{S_LSTM} & - & - & 61.6 & 35.3 \\
		Gated S-CNN~\citep{Gated_SCNN} & 65.8 & 39.5 & 76.0 & 48.4 \\
		\hline
		\bf IRS~(WHOS) & 55.2 & 27.5 & 60.3 & 33.5 \\
		\bf IRS~(LOMO) & 57.7 & 29.0 & 68.0 & 37.8 \\ 
		\bf IRS~(CNN) & \bf 72.7 & \bf 48.1 & \bf 80.2  & \bf 58.5  \\  %
		\hline 
		\hline
		SCSP$^*$ \citep{chen2016similarity} & 51.9 & 26.4 & - & - \\
		NFST$^*$ \citep{zhang2016learning} & 61.0 & 35.7 & 71.6 & 46.0 \\
		\hline
		\bf IRS~(WHOS+LOMO+CNN)$^*$ & \bf 73.9 & \bf 49.4 & \bf  81.4 & \bf 59.9 \\ \whline
	\end{tabular}
	\label{tab:CMC_conventional_market}
\end{table}

\begin{table*} 
	\centering
	\renewcommand{\arraystretch}{1}
	\setlength{\tabcolsep}{0.075cm}
	\caption{Effects of embedding space on re-id performance
		in our proposed IRS model. The LOMO visual feature
		were used on all datasets. 
		We adopted the 485/486 people split on CUHK01 and 
		the manually labelled person images on CUHK03.
		SQ: Single-Query; MQ: Multi-Query.
	}
	\vskip -10pt 
	\begin{tabular}{|c|cccc|cccc|cccc|cc|cc|}
		\whline
		Dataset  & 
		\multicolumn{4}{c|}{VIPeR} & 
		\multicolumn{4}{c|}{CUHK01} & 
		\multicolumn{4}{c|}{CUHK03} & 
		\multicolumn{4}{c|}{Market-1501} \\ \hline 
		Rank (\%)    
		& R1 & R5 & R10 & R20 
		& R1 & R5 & R10 & R20
		& R1 & R5 & R10 & R20
		& R1(SQ) & mAP(SQ) & R1(MQ) & mAP(MQ)  \\ \hline \hline
		OneHot Feature Coding
		& {\bf 45.1} & {\bf 74.6} & {\bf 85.1}  & {\bf 93.3} 
		& {\bf 68.3} & {\bf 86.7} & {\bf 92.6}  & {\bf 96.2}  
		& {\bf 61.6} & {\bf 87.0} & 94.6  & {\bf 98.0} 
		& {\bf 57.7} & {\bf 29.0} & {\bf 68.0} & {\bf 37.8} \\   \hline
		FDA Feature Coding 
		& {\bf 45.1} & {\bf 74.6} & {\bf 85.1} & {\bf 93.3} 
		& {\bf 68.3} & {\bf 86.7} & {\bf 92.6} & {\bf 96.2} 
		& 60.8 & {\bf 87.0} & {\bf 95.2} & 97.7 & 55.6   & 27.5   & 67.5  & 36.8  \\  %
		\hline
		Random Feature Coding 
		& 44.8 & 73.4 & 84.8 & 92.7 & 61.3 & 83.4 & 89.5  & 94.2 & 51.7   & 79.4   & 87.4  & 93.0& 47.4   & 21.1   & 48.5  & 23.2  \\ 
		\whline
	\end{tabular}
	\label{tab:CMC_TISP}
\end{table*}

\vspace{0.1cm}
\noindent {\bf\em Evaluation on CUHK01. } 
Table \ref{tab:CMC_conventional_CUHK01} shows a comprehensive
comparison of the IRS model with existing competitive re-id models on the CUHK01
benchmark~\citep{transferREID}.
It is clear that the proposed IRS model achieves 
the best re-id accuracy under both training/test split protocols.
Note that, HER \citep{wang2016highly} is IRS-FDA(LOMO).
Specifically, for the 486/485 split, our IRS(CNN) method 
surpassed the deep learning DGD model \citep{xiao2016learning}, the
second best in this comparison, 
by Rank-1 $2.0\%(68.6\!-\!66.6)$. 
For the 871/100 split, IRS(CNN) yields a greater performance boost
over DGD with improvement on Rank-1 at $12.6\%(84.4\!-\!71.8)$.
It is also worth pointing out that the DGD model was trained using
data from other 6 more datasets and further carefully fine-tuned on CUHK01. In
contrast, our IRS(CNN) model was only trained on CUHK03 without
fine-tuning on CUHK01, and the CNN architecture we adopted closely
resembles to that of DGD. 
By fusing multiple features, the performance margin of IRS(WHOS+LOMO+CNN) over the
existing models is further enlarged under both splits, achieving Rank-1
$11.7\%(80.8\!-\!69.1)$ boost over NFST~\citep{zhang2016learning} and
Rank-1 $16.6\%(88.4\!-\!71.8)$ boost over SICI~\citep{wangjoint}, respectively. 
Compared to VIPeR, the overall re-id performance
advantage of the IRS model on CUHK01 is greater over existing
models. This is due to not only identity prototype regression based
feature embedding, but also less domain drift from CUHK03 to
CUHK01, given that the CNN feature used by IRS was trained on 
CUHK03.

\vspace{0.1cm}
\noindent {\bf\em Evaluation on CUHK03. }   
The person re-id performance of different methods as compared to
the IRS model on CUHK03 \citep{Li_DeepReID_2014b} is reported in Table \ref{tab:CMC_conventional_CUHK03}.
We tested on both the manually labelled and automatically detected bounding boxes.
Similar to VIPeR and CUHK01, our IRS model surpassed clearly all compared methods 
in either single- or multi-feature setting given
manually labelled bounding boxes. Importantly, this advantage remains
when more challenging detected bounding boxes were used, whilst other
strong models such as NFST and GOG suffered more significant
performance degradation. 
This shows both the robustness of our IRS model against misalignment
and its greater scalability to real-world deployments.


\vspace{0.1cm}
\noindent {\bf\em Evaluation on Market-1501. }
We evaluated the re-id performance of existing models against the
proposed IRS model on the Market-1501 benchmark \citep{zheng2015scalable}.
The bounding boxes of all person images of this dataset were generated 
by an automatic pedestrian detector. Hence, this dataset presents a more
realistic challenge to re-id models than conventional re-id datasets
with manually labelled bounding boxes.
Table \ref{tab:CMC_conventional_market} shows 
the clear superiority of our IRS model over all competitors. 
In particular, our IRS model achieved Rank-1 $73.9\%$ for single-query and
Rank-1 $81.4\%$ for multi-query, significantly better than 
the strongest alternative method, the deep Gated S-CNN
model~\citep{Gated_SCNN}, by $8.1\%(73.9\!-\!65.8)$ (single-query) and
$5.4\%(81.4\!-\!76.0)$ (multi-query). 
Similar advantages hold when compared using the mAP metric.

In summary, these comparative evaluations on the performance of batch-wise re-id model
learning show that the IRS model outperforms
comprehensively a wide range of existing re-id methods including both
hand-crafted and deep learning based models. 
This validates the effectiveness and advantages of learning a re-id
discriminative feature embedding using the proposed approach on identity
regression.
%

\begin{table} [h]
	\centering
	\renewcommand{\arraystretch}{0.92}
	\setlength{\tabcolsep}{0.2cm}
	\caption{
		Effects of feature choice in re-id performance using
		the IRS model with OneHot Feature Coding.
	}
	\vskip -10pt
	\begin{tabular}{|c|cccc|}
		\whline
		Dataset  & \multicolumn{4}{c|}{VIPeR} \\ \hline 
		Rank (\%)     & R1   & R5   & R10   & R20    \\ \hline
		WHOS \citep{lisanti2015person} & 44.5& \bf 75.0 & \bf 86.3 & \bf 93.6\\
		LOMO \citep{liao2015person} & \bf 45.1 & 74.6 & 85.1 &  93.3 \\ 
		CNN \citep{xiao2016learning} & 33.1 &  59.9 &71.5  & 82.2  \\ 
		\hline  
		WHOS+LOMO & 53.0 & 79.8 & 89.6 & 95.5\\ 
		CNN+LOMO & 49.9 & 77.5 &86.9 & 93.8 \\
		WHOS+CNN & 49.7 & 78.0 & 87.9 & 94.4\\ 
		WHOS+LOMO+CNN & \bf 54.6 & \bf 81.5 & \bf 90.3 &  \bf 95.7\\
		\hline \hline

		Dataset  & \multicolumn{4}{c|}{CUHK01 (486/485 split)} \\ \hline 
		WHOS \citep{lisanti2015person} & 48.8 & 73.4 & 81.1 & 88.3\\
		LOMO \citep{liao2015person} & 68.3 & 86.7 &92.6  & 96.2 \\ 
		CNN \citep{xiao2016learning} & \bf 68.6 & \bf 89.3 & \bf 93.9 & \bf 97.2  \\  
		\hline
		WHOS+LOMO & 71.2 & 90.0 & 94.4 & 97.3 \\
		CNN+LOMO &  79.8 & 93.6 &  96.3 &  98.2 \\ 
		WHOS+CNN & 76.1 & 92.9 & 96.1 & 98.2 \\
		WHOS+LOMO+CNN & \bf 80.8 & \bf 94.6 & \bf 96.9 & \bf 98.7\\
		\hline
		\hline
		Dataset  & \multicolumn{4}{c|}{CUHK01 (871/100 split)}  \\ \hline 
		WHOS \citep{lisanti2015person} & 77.0 & 92.8 & 96.5 & 99.2\\
		LOMO \citep{liao2015person} & 80.3 & 94.2 & 96.9  & 99.5 \\ 
		CNN \citep{xiao2016learning} & \bf 84.4 & \bf 98.2 & \bf 99.8 & \bf 100  \\  %
		\hline
		WHOS+LOMO & 83.6 & 95.4 & 98.8 &\bf  100 \\
		CNN+LOMO & 88.0 & 98.3  & 99.5 &\bf  100 \\ 
		WHOS+CNN & \bf 89.0 & 98.5 & \bf 99.6 & \bf 100 \\
		WHOS+LOMO+CNN & 88.4 & \bf 98.8 & \bf 99.6 & \bf 100\\
		\hline \hline

		Dataset  & \multicolumn{4}{c|}{CUHK03 (Manually)} \\ \hline 
		WHOS \citep{lisanti2015person} & 59.6 & 87.2 & 92.8 & 96.9\\
		LOMO \citep{liao2015person} & 61.6 & 87.0 & 94.6  & 98.0 \\ 
		CNN \citep{xiao2016learning} & \bf 81.5 & \bf 95.7 & \bf 97.1  & \bf 98.0  \\  
		\hline
		WHOS+LOMO & 65.2 & 92.2 & 96.8 & \bf 99.1\\
		CNN+LOMO & \bf 82.6 & 96.0 & 97.5 & 98.6 \\
		WHOS+CNN & 80.4 & 95.7 & 98.0 & 98.4\\
		WHOS+LOMO+CNN & 81.9 & \bf 96.5 & \bf 98.2 & 98.9\\
		\hline \hline
		Dataset  & \multicolumn{4}{c|}{CUHK03 (Detected)} \\ \hline
		WHOS \citep{lisanti2015person} & 50.6 & 82.1 & 90.4 & 96.1\\
		LOMO \citep{liao2015person} & 53.4 & 83.1 &   91.2 & 96.4 \\ 
		CNN \citep{xiao2016learning} & \bf 80.3 & \bf 96.3 & \bf 98.6  & \bf 99.0  \\ 
		\hline  
		WHOS+LOMO& 59.9 & 89.4 & 95.5 & 98.5\\
		CNN+LOMO & 82.4  & 95.7 & 97.4 & 98.4 \\
		WHOS+CNN & 81.1 & 95.4 & 97.5 & \bf 98.6\\
		WHOS+LOMO+CNN & \bf 83.3 & \bf 96.2 & \bf 97.9 & \bf 98.6 \\
		\hline \hline

		Dataset  & \multicolumn{4}{c|}{Market-1501} \\ \hline 
		Query Per Person   & \multicolumn{2}{c|}{Single-Query}  & \multicolumn{2}{c|}{Multi-Query}    
		\\ \hline
		Metric (\%)   & R1   & mAP   & R1   & mAP    \\ \hline
		WHOS \citep{lisanti2015person} & 55.2 & 27.5 & 60.3 & 33.5\\
		LOMO \citep{liao2015person} & 57.7 & 29.0 & 68.0 & 37.8 \\ 
		CNN \citep{xiao2016learning} & \bf 72.7 & \bf 48.1 & \bf 80.2  & \bf 58.5  \\ 
		\hline
		WHOS+LOMO & 62.4 & 33.6 & 69.0 & 41.0\\
		CNN+LOMO &  73.0 & 48.5 &  80.9 & 59.1 \\
		WHOS+CNN & 72.8 & 48.3 & 80.3 & 58.7\\
		WHOS+LOMO+CNN & \bf 73.9 & \bf 49.4 & \bf 81.4 & \bf 59.9\\
		\whline
	\end{tabular}
	\label{tab:vfeat}
\end{table}

\vspace{0.1cm}
\noindent \textbf{(II) Effects of Embedding Space Design. } 
To give more insight on why and how the IRS model works, we evaluated
the effects of embedding space design in our IRS model. 
To this end, we compared the three coding methods as described in
Sec. \ref{sec:regression}: {\em OneHot Feature Coding} in the proposed
{\em Identity Regression Space}, {\em FDA Feature Coding} by
\citet{wang2016highly}, and {\em Random Feature Coding} by \citet{zhu2016heat}.
%
In this experiment, we used 
the LOMO feature on all four datasets,
the 485/486 people split on CUHK01,
and the manually labelled bounding boxes on CUHK03.
For Random Coding, we performed 10 times and used the averaged results
to compare with the OneHot Feature Coding and the FDA Feature Coding.
The results are presented in Table \ref{tab:CMC_TISP}. 
We have the following observations: 

(i) The embedding space choice plays a clear role in IRS re-id model learning and 
a more ``semantic'' aligned (both OneHot and FDA) coding has the
advantage for learning a more discriminative IRS re-id model.
One plausible reason is that the Random Coding may increase the model
learning difficulty resulting in an inferior feature embedding,
especially given the small sample size nature of re-id model
learning. 
Instead, by explicitly assigning identity class ``semantics''
(prototypes) to individual dimensions of the embedding space, the
feature embedding learning is made more selective and easier to optimise. 

(ii) Both the OneHot and FDA Feature Coding methods yield the same re-id
accuracy on both VIPeR and CUHK01.
This is because on either dataset each training identity has the same number of images 
($2$ for VIPeR and $4$ for CUHK01), under which the FDA Coding (Eq. \eqref{eq:FDA_Y}) 
is equivalent to the OneHot Feature Coding (Eq.~\eqref{eq:onehot_Y}). 
%
%

(iii) Given the different image samples available per training person identity on CUHK03 and Market-1501,
FDA Coding is slightly inferior to OneHot Feature Coding.
This is interesting given the robust performance of FDA on
conventional classification problems. Our explanation is rather
straightforward if one considers the unique characteristics of the
re-id problem where the training and test classes are {\em completely}
non-overlapping. That is, 
the test classes have no training image samples. 
In essence, 
{the re-id problem is conceptually similar to the
problem of Zero-Shot Learning (ZSL)}, in contrast to the conventional
classification problems where test classes are sufficiently
represented by the training data, i.e. totally overlapping.
More specifically, learning by the FDA criterion optimises a model to
the training identity classes given sufficient samples per class but
it does not work well with small sample sizes, and more critically, it
does {\em not necessarily} optimise the model for previously unseen
test identity classes. This is because if the training identity
population is relatively small, as in most re-id datasets, an unseen
test person may not be similar to any of training people, 
That is, the distributions of 
the training and test population may differ significantly. 
Without any prior knowledge, a good representation of an unseen test
class is some unique combination of all training persons {\em
  uniformly} without preference. Therefore, a feature embedding
optimised uniformly without bias/weighting by the training class data
sampling distribution is more likely to better cope with more diverse
and unseen test classes, by better preserving class diversity in the
training data {\em especially given the small sample size challenge}
in re-id training data. This can be seen from the regularised
properties of the OneHot Feature Coding in Sec. \ref{sec:HER}.

\vspace{0.1cm}
\noindent \textbf{(III) Effect of Features. } 
We evaluated three different features (WHOS,
LOMO, and CNN) individually and also their combinations used in our
IRS model with the OneHot Feature Coding in Table \ref{tab:vfeat}.
When a single type of feature is used, it is found that CNN feature 
is the best except on VIPeR, and LOMO is more
discriminative than WHOS in most cases. 
The advantage of CNN feature over hand-crafted LOMO
and WHOS is significant given larger training data in CUHK03 and Market-1501, 
yielding a {\em gain} of $19.9\%$ (CUHK03 (Manual)), $26.9\%$
(CUHK03 (Detected)), and $15.0\%$ (Market-1501) over LOMO
in Rank-1.
Without fine-tuning a CUHK03 trained model on the target
domains, CNN feature still performs the best on CUHK01 due to the high similarity in
view conditions between CUHK01 and CUHK03. 
CNN feature performs less well on
VIPeR due to higher discrepancy in view conditions between
VIPeR and CUHK03, i.e. the domain shift problem
\citep{Transfer_ICCV13,pan2010survey}.

We further evaluated multi-feature based performance
by score-level fusion. It is evident that most combinations lead to
improved re-id accuracy,
and fusing all three features often generate the best results.
This confirms the previous findings that 
different appearance information can be encoded by distinct features 
and their fusion enhances re-id matching 
\citep{Anton_2015_CoRR,zhang2016learning,GOG,chen2016similarity}.

\changed{
\begin{table}[h]
	\centering
	\setlength{\tabcolsep}{0.3cm}
	\caption{
		Evaluation on deep learning regression (DLR)
			on CUHK03 (Manually).
			Deep model: DGD \citep{xiao2016learning}. 
			DLR$^\text{$n$-FC}$: $n\in\{1,2,3\}$ FC layers added in DLR. 
			RR = Ridge Regression.
	}
	\vskip -0.3cm
	\begin{tabular}{|c|cccc|}
		\whline
		Rank (\%) & R1 & R5 & R10 & R20 \\ \hline \hline
		CNN Feature  & 73.7 & 91.5 & 95.0 & 97.2\\ 
		Softmax Prediction  & 73.3 & 91.0 & 93.9 & 96.4 \\ 
		\hline
		{\bf IRS}(DLR$^\text{1-FC}$)
		& 75.1 & 92.7 & 95.3 & 97.5 \\  
		{\bf IRS}(DLR$^\text{2-FC}$)
		& 76.6 & 93.1 & 95.9 & \bf 98.1 \\
		{\bf IRS}(DLR$^\text{3-FC}$)
		& 74.2 & 92.5 & 94.8 & 97.1 \\ \hline 
		CNN + {\bf IRS}(RR) & \bf 81.5 & \bf 95.7 & \bf 97.1 &  98.0\\
		\whline
	\end{tabular}
	\label{tab:deep_softmax}
\end{table}
}

%

\vspace{0.1cm}
\noindent \changed{
\textbf{(IV) Deep Learning Regression. }
%
Apart from the Ridge
Regression (RR) algorithm \citep{hoerl1970ridge,zhang2010regularized}, the IRS concept
can be also realised in deep learning,
i.e. Deep Learning Regression (DLR).
We call this IRS implementation as {\bf IRS(DLR)}.
For this experiment, we adopted the DGD CNN model \citep{xiao2016learning}
and the CUHK03 (Manual) dataset.
In training IRS(DLR), we first trained the DGD to convergence with the softmax cross-entropy loss.
Then, we added $n$(=1,2,3) new 512-dim FC layers (including ReLU activation) with random parameter initialisation on top of DGD.
Finally, we frozen all original DGD layers and optimised the new layers only
by $L_2$ loss.

In this test, we compared with 
the DGD (1) CNN Features and
(2) Softmax Predictions (considered as some sort of IRS features
although not strictly the same due to different modelling designs).
%
%
We observed in Table \ref{tab:deep_softmax} that:
(1) IRS(DLR) outperforms both CNN Features and Softmax Prediction. 
This indicates the benefit of IRS in a deep learning context.
(2) IRS(DLR) is relatively inferior to CNN+IRS(RR),
suggesting that a deep learning model is not necessarily superior in regressing IRS
when given limited training data. Moreover,
IRS(RR) is superior on model learning efficiency,
hence more suitable for incremental model update.
}

\begin{table} [h]
	\centering
	\setlength{\tabcolsep}{0.3cm}
	\caption{
		Evaluation on the complementary effect of deep model pre-training based transfer learning (TL) and IRS
		on VIPeR. Deep model: DGD \citep{xiao2016learning}. 
		$*$: Reported result in \citep{xiao2016learning}. 
	}
	\vskip -0.3cm
	\begin{tabular}{|c|cccc|}
		\whline
		Rank (\%)     & R1   & R5   & R10   & R20    \\ \hline \hline
		W/O TL$^*$
		& 12.3 & - & - & -\\
		{W} TL & 34.1 & 66.3 & 76.2 &  83.7 \\
		\hline 
		TL + {\bf IRS}(RR) & \bf 39.9 & \bf 70.6 & \bf 79.3 &  \bf 86.2\\
		\whline
	\end{tabular}
	\label{tab:TL}
\end{table}

\vspace{0.1cm}
\noindent \changed{
\textbf{(V) Complementary of Transfer Learning and IRS. }
Transfer learning (TL) is another {independent} scheme for solving the SSS problem.
We tested the benefit of deep learning pre-trained TL and IRS.
We evaluated three methods based on the DGD \cite{xiao2016learning}: 
(1) \textbf{\em W/O TL}: Trained the DGD on VIPeR training data (632 images) only. 
(2) \textbf{\em W TL}: First pre-trained the DGD on 26,246 CUHK03 images for knowledge transfer learning, 
then fine-tuned on the VIPeR training data.
(3) \textbf{\em TL + IRS(RR)}: First adopted the CUHK03 pre-trained and VIPeR fine-tuned DGD to extract CNN features,
then deployed the ridge regression based IRS to train the final re-id feature embedding model.
All three models were evaluated on the same VIPeR test data.
Table \ref{tab:TL} shows that:
(1) Pre-training based TL significantly improves re-id performance.
This demonstrates the benefit of TL in solving the SSS problem.
(2) IRS clearly further improves the re-id accuracy.
This verifies the additional benefits of IRS and 
the complementary advantage of TL and IRS
to a deep learning model for solving the SSS challenge.
}

\begin{table}[h]
	\centering
	\renewcommand{\arraystretch}{1}
	\setlength{\tabcolsep}{0.2cm}
	\caption{
		\eddy{Comparing subspace learning models with different features.}
	}
	\vskip -10pt
	\begin{tabular}{|c|cccc|}
		\whline
		Dataset - Feature & \multicolumn{4}{c|}{VIPeR - WHOS} \\ \hline 
		Rank (\%)     & R1   & R5   & R10   & R20    \\ \hline
		KISSME \citep{KISSME_CVPR12} & 28.7 & 57.2 & 72.6 & 86.1 \\
		kLFDA \citep{xiong2014person} & 40.1 & 68.5 & 81.2 & 91.7 \\
		XQDA \citep{liao2015person}  & 35.1 & 63.9 & 74.9 & 86.0\\
		NFST~\citep{zhang2016learning} & 43.6 & 74.1& 86.1 & 92.7\\
		\hline  
		IRS & \bf 44.5 & \bf 75.0  & \bf 86.3 &  \bf 93.6\\
		\hline 
		\hline
		Dataset - Feature & \multicolumn{4}{c|}{VIPeR - LOMO} \\ \hline 
		KISSME \citep{KISSME_CVPR12} & 22.1 & 53.4 & 68.8 & 83.8 \\
		kLFDA \citep{xiong2014person} & 38.6 & 69.2& 80.4 & 89.2 \\
		XQDA \citep{liao2015person}  & 40.0 & 68.1 & 80.5 & 91.1\\
		NFST~\citep{zhang2016learning} & 42.3 & 71.5 & 82.9 & 92.1\\
		\hline  
		IRS & \bf 45.1 & \bf 74.6 & \bf 85.1 &  \bf 93.3\\
		\hline 
		\hline
		Dataset - Feature & \multicolumn{4}{c|}{VIPeR - CNN} \\ \hline 
		KISSME \citep{KISSME_CVPR12} & 22.6 & 46.9 & 59.0 & 72.7 \\
		kLFDA \citep{xiong2014person} & 30.9 & 55.6 & 65.7 & 75.0 \\
		XQDA \citep{liao2015person}  & 11.7 & 26.2 & 35.5 & 48.1\\
		NFST~\citep{zhang2016learning} & 31.2 & 56.0 & 67.2 & 78.4\\
		\hline  
		IRS & \bf 33.1 & \bf 59.9 & \bf 71.5 &  \bf 82.2\\
		\hline 
		\hline
		
		Dataset - Feature & \multicolumn{4}{c|}{CUHK03(M) - WHOS} \\ \hline 
		Rank (\%)     & R1   & R5   & R10   & R20    \\ \hline
		KISSME \citep{KISSME_CVPR12} & 31.6 & 63.4 & 76.6 & 88.3 \\
		kLFDA \citep{xiong2014person} & 32.9 & 59.2 & 75.7 & 82.6 \\
		XQDA \citep{liao2015person}  & 41.1 & 66.5 & 77.2 & 86.6 \\
		NFST~\citep{zhang2016learning} & 34.4 & 59.7 & 68.2 & 77.6\\
		\hline  
		IRS & \bf 59.6 & \bf 87.2 & \bf 92.8 &  \bf 96.9\\
		\hline 
		\hline
		Dataset - Feature & \multicolumn{4}{c|}{CUHK03(M) - LOMO} \\ \hline 
		KISSME \citep{KISSME_CVPR12} & 32.7 & 68.0 & 81.3 & 91.4 \\
		kLFDA \citep{xiong2014person} & 45.8 & 77.1 & 86.8 & 93.1 \\
		XQDA \citep{liao2015person}  & 52.2 & 82.2 & 92.1 & 96.3\\
		NFST~\citep{zhang2016learning} & 58.9 & 85.6 & 92.5 & 96.3\\
		\hline  
		IRS & \bf 61.6 & \bf 87.0& \bf 94.6 &  \bf 98.0\\
		\hline 
		\hline
		Dataset - Feature & \multicolumn{4}{c|}{CUHK03(M) - CNN} \\ \hline 
		KISSME \citep{KISSME_CVPR12} & 73.8 & 94.0 & 96.2 & \bf 98.0 \\
		kLFDA \citep{xiong2014person} & 76.0 & 92.3 & 96.0 & \bf 98.0 \\
		XQDA \citep{liao2015person}  & 70.8 & 92.0 & 96.2 & 97.9\\
		NFST~\citep{zhang2016learning} & 62.6 & 78.9 &85.5 & 89.7\\
		\hline  
		IRS & \bf 81.5& \bf 95.7 & \bf 97.1 &  \bf 98.0\\
		\whline
	\end{tabular}
	\label{tab:subspace}
\end{table}

\vspace{0.1cm}
\noindent \sgg{
\textbf{(VI) Comparisons to Subspace/Metric Learning Models. }
We performed comparative experiments on
four subspace and metric learning models including
KISSME~\citep{KISSME_CVPR12}, kLFDA~\citep{xiong2014person},
XQDA~\citep{liao2015person}, and NFST~\citep{zhang2016learning},
using three different types of features (WHOS, LOMO, CNN) and identical 
training/test data.
We utilised the same subspace dimension for XQDA and our IRS,
i.e. the number of training person classes.
We conducted this evaluation on VIReR and CUHK03 (Manual).
Table \ref{tab:subspace} shows that the proposed IRS model 
consistently surpasses all the compared alternative models.
This again suggests the advantages of IRS in 
learning discriminative re-id models.
}

\begin{figure}
	\centering
	\includegraphics[width=0.98\linewidth]{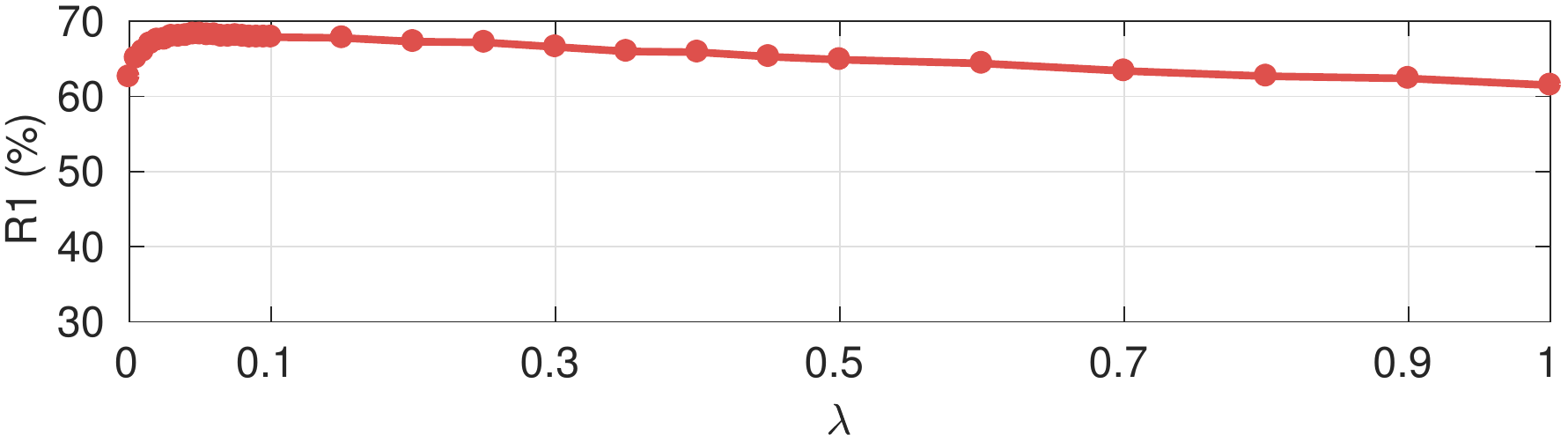} \\
	\includegraphics[width=0.98\linewidth]{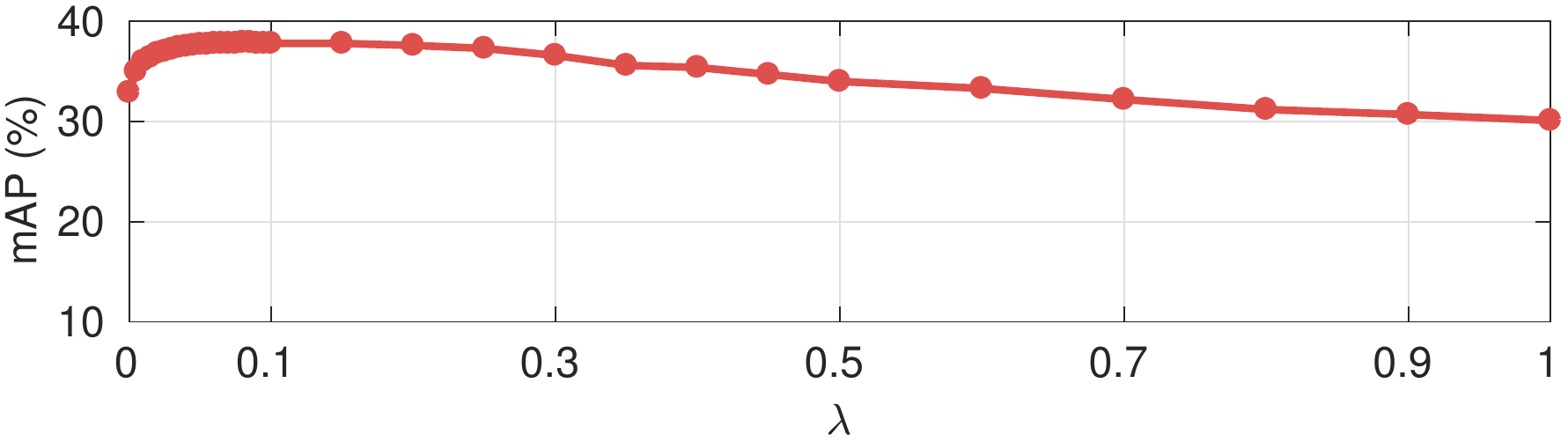}
	\vskip -12pt 
	\caption{
	\eddy{Regularisation sensitivity on the Market-1501 dataset.
          The multi-query setting was used.
      }
	}
	\label{fig:regularisation}
\end{figure}

\vspace{0.1cm}
\noindent \textbf{(VII) Regularisation Sensitivity. } 
We analysed the sensitivity of the only free parameter $\lambda$ in Eq. \eqref{eq:HER_solution} 
which controls the regularisation strength of our IRS model. 
This evaluation was conducted with the LOMO feature in the multi-query setting 
on Market-1501 \citep{zheng2015scalable}. 
\eddy{Specifically, we 
evaluated the Rank-1 and mAP performance with
$\lambda$ varying from $0$ to $1$. 
Fig. \ref{fig:regularisation} shows that 
our IRS model has a large satisfactory range of $\lambda$
and therefore not sensitive.
We set $\lambda = 0.1$ in all evaluations.}

\begin{table}[h!]
	\renewcommand{\arraystretch}{1}
	\setlength{\tabcolsep}{0.2cm}
	\caption{
		Model complexity and training costs of person re-id models. 
		{\em Metric}: Model training time (in seconds), smaller is better.
		%
	}
	\vskip -10pt
	\begin{tabular}{|c|c|c|c|c|}
		\swhline
		Dataset & VIPeR & CUHK01 & CUHK03 & Market-1501  \\ \hline 
		Training Size & 632 & 1940 & 12197 & 12936 \\ \hline \hline
		MLAPG & 50.9 & 746.6  &  4.0$\times 10^4$ & - \\
		kLFDA  & 5.0 & 45.9 & 2203.2   &  1465.8   \\
		XQDA   &  4.1 &  51.9 &  3416.0   &   3233.8      \\ 
		NFST  &  1.3 &  6.0 & 1135.1  &   801.8 \\ 
		\hline 
		\bf IRS  & \bf 1.2 & \bf 4.2 & \bf 248.8 & \bf 266.3 \\\whline
	\end{tabular}
	\label{tab:model_comlexity}
\end{table}

\vspace{0.2cm}
\noindent \textbf{(VIII) Model Complexity. }
In addition to model re-id accuracy, we also examined the model
complexity and computational costs, in particular model training time.
We carried out this evaluation by comparing our IRS model with
some strong metric learning methods including kLFDA \citep{xiong2014person}, 
XQDA \citep{liao2015person}, 
MLAPG \citep{Liao_2015_ICCV}, 
and NFST \citep{zhang2016learning}.
Given $n$ training samples represented by $d$-dimensional feature vectors,
it requires $\frac{3}{2}dnm + \frac{9}{2}m^3$ ($m = \min(d,n)$) 
floating point addition and multiplications \citep{penrose1955generalized}
to perform an eigen-decomposition for solving either a generalised eigen-problem
\citep{xiong2014person,liao2015person} or a null space \citep{zhang2016learning},
whereas solving the linear system of the IRS model (Eq.~\eqref{eq:HER_solution})
takes $\frac{1}{2}dnm + \frac{1}{6}m^3$~\citep{cai2008srda}.
Deep learning models \citep{Ahmed2015CVPR,xiao2016learning,Gated_SCNN}
are not explicitly evaluated since they  
are usually much more demanding in computational overhead,
requiring much more training time (days or even weeks) and more
powerful hardware (GPU). In this evaluation, we adopted the LOMO
feature for all datasets and all the models compared, the 485/486
people split on CUHK01, the manually labelled person bounding boxes on CUHK03,
and the single-query setting on Market-1501.

For each model, we recorded and compared 
the average training time of 10 trials performed
on a 
workstation with $2.6$GHz CPU. 
Table~\ref{tab:model_comlexity} presents the training time of
different models (in seconds).
On the smaller VIPeR dataset, our IRS model training needed only
$1.2$ seconds, similar as NFST 
and $42.4$ times faster than MLAPG.  
On larger datasets CUHK01, CUHK03 and Market-1501,
all models took longer time to train and training the IRS model
remains the fastest with speed-up over MLAPG 
enlarged to $177.8$ / $160.8$ times 
on CUHK01 / CUHK03, respectively\footnote{The MLAPG model failed to converge on Market-1501.}.
This demonstrates the advantage of 
the proposed IRS model over existing competitors for scaling up to
large sized training data.

%
%


\begin{table*}[t!]
	\centering
	\renewcommand{\arraystretch}{1}
	\setlength{\tabcolsep}{0.085cm}
	\caption{
		Comparing passive Incremental Learning (IL) vs.
		Batch-wise Learning (BL) using the IRS model.
		ALT: Accumulated Learning Time,
		i.e. the summed time for training all the 151 IRS models when 
		the label number is increased from 50 to 200 one by one.
	}
	\vskip -10pt
	\begin{tabular}{|c|c|cccc|c|cccc|c|cccc|c|cccc|c|}
		\whline
		\multicolumn{2}{|c|}{Dataset} & \multicolumn{5}{c|}{VIPeR} & \multicolumn{5}{c|}{CUHK01} & \multicolumn{5}{c|}{CUHK03} & \multicolumn{5}{c|}{Market-1501} \\ \hline
		\multicolumn{2}{|c|}{Label \#} & 50   & 100   & 150   & 200 & {\em ALT}  
		& 50   & 100   & 150   & 200 & {\em ALT} 
		& 50   & 100   & 150   & 200 & {\em ALT}  
		& 50   & 100   & 150   & 200 & {\em ALT}  \\ \hline \hline
		{Time} & BL & 0.23 & 0.23 & 0.25 & 0.26 & 36.5  & 1.43 & 1.51 & 1.57  & 1.66 & 232.8 & 20.4 & 21.7 & 22.4 & 24.5  & 3349.9 & 119.5 & 121.5 & 125.6 & 140.3 & $1.9 \times 10^4$
		\\
		(sec.)& IL & \bf 0.02 & \bf 0.02 &\bf 0.02 &\bf 0.03 &\bf 3.28 &\bf 0.14 &\bf 0.15 &\bf 0.16  &\bf 0.17 &\bf 23.4 &\bf 1.62 &\bf 1.69 &\bf 1.70 &\bf 1.81  &\bf 257.0 &\bf 1.94 &\bf 5.05&\bf 6.61 &\bf 9.60  &\bf 877.3 
		\\ 
		\hline
		{R1} & BL &\bf 20.6 &\bf 29.2 &\bf 34.9 &\bf 38.9 & - &\bf 21.9 &\bf 37.3 &\bf 46.5  &\bf 52.5 & - &\bf 24.0 &\bf 35.2 &\bf 40.5  &\bf 43.8 &- &\bf 28.6 &\bf 44.5  &\bf 51.7  &\bf 55.2 & -
		\\
		(\%) & IL & 19.4 &\bf 29.2 & 33.6 & 37.2 & - & 20.8 & 35.6 & 45.3  & 51.5 & - &  22.1 & 33.0  &  38.8 & 41.7 & - & 27.5  & 44.2  & 50.6   & 54.3  & -\\
		\whline
	\end{tabular}
	\label{tab:incremental}
\end{table*}

\begin{table*}[t]
	\centering
	\renewcommand{\arraystretch}{1}
	\setlength{\tabcolsep}{0.16cm}
	\caption{
		Evaluation on the active incremental learning algorithm.
		{\em Metric}: Rank-1 rate (\%).
	}
	\vskip -10pt
	\label{tab:CMC_active}
	\begin{tabular}{|c|cccc|cccc|cccc|cccc|}
		\whline
		{Dataset}  & \multicolumn{4}{c|}{VIPeR} & \multicolumn{4}{c|}{CUHK01} & \multicolumn{4}{c|}{CUHK03} & \multicolumn{4}{c|}{Market-1501} \\ 
		\hline
		{Label \#} 
		& 50   & 100   & 150  & 200 
		&  50   & 100   & 150 & 200 
		&  50   & 100   & 150 & 200 
		&  50   & 100   & 150 & 200\\ 
		\hline
		\hline
		Random  
		& 19.4 & 29.2 & 33.6 & 37.2  & 20.8 & 35.6 & 45.3  & 51.5 &  22.1 & 33.0  &  38.8 & 41.7 & 27.5 & 44.2 & 50.6 & 54.3 \\
		Density~\citep{ebert2012ralf} 
		& 18.4 & 26.8  &  33.5 & 37.5 & 23.3 & 37.0 & 44.5  &50.0  & 23.7 & 34.8 &  40.2 & 42.7 & 32.3 & 46.2 & 51.5 & 53.9\\
		\bf Joint$\mathbf{E}^2$ 
		& \bf 23.4 & \bf 31.4 & \bf 36.5 & \bf  40.9 &  \bf  29.9  & \bf   39.7 & \bf 47.1 & \bf 52.2 & \bf 25.1 & \bf 36.8 & \bf 41.3 & \bf 43.0 & \bf 36.5 & \bf 50.7 & \bf 54.8 & \bf 58.2\\ \whline
		
	\end{tabular}
\end{table*}

\subsection{Incremental Person Re-Id Evaluation}
\label{sec:eval_incre}

We further evaluated the performance of our IRS model using the
incremental learning IRS$^\text{inc}$ algorithm (Sec. \ref{sec:incHER}). 
This setting starts with a small number, e.g. $10$ of labelled true match training pairs,
rather than assuming a large pre-collected training set.
Often, no large sized labelled data is available in typical deployments
at varying scenes in advance.
More labelled data will arrive one by one over time during deployment
due to human-in-the-loop verification.
In such a setting, a re-id model can naturally evolve through deployment life-cycle and 
efficiently adapt to each application test domain. 
In this context, we consider two incremental re-id model learning scenarios:
{\bf (I)} {\em Passive} incremental learning where unlabelled person images are 
randomly selected for human to verify;
{\bf (II)} {\em Active} incremental learning where person images are actively
determined by the proposed JointE$^2$ active learning algorithm (Sec. \ref{sec:active}).

\vspace{0.1cm}
\noindent {\bf Incremental Re-Id Evaluation Protocol.} 
Due to the lack of access to large sized training samples in batch,
incrementally learned models are typically less powerful
than batch learned models \citep{poggio2001incremental,ristin2014incremental}.
Therefore, it is critical to evaluate how much performance drop is introduced
by the Incremental Learning (IL) algorithm, IRS$^\text{inc}$, as compared to 
the corresponding Batch-wise Learning (BL) and how much efficiency is
gained by IL.

We started with $10$ labelled identities, i.e. cross-camera truth matches
of $10$ persons, and set the total labelling budget to $200$ persons.
For simplicity, we selected four test cases
with $50,100,150,200$ labelled identities respectively and evaluated
their model accuracy and training cost.
To compare the Accumulated Learning Time (ALT)\footnote{
The BL model needs to be trained once only after all 200 person classes are labelled
when we consider the batch-wise model learning setting (Sec. \ref{sec:eval_batch}). 
However, here we consider instead the incremental learning setting
with the aim to evaluate the proposed incremental learning algorithm 
in both training efficiency and effectiveness,
as compared to the batch learning counterpart when 
deployed for model incremental update.
Given the batch-wise learning strategy, 
incremental model update can only be achieved by
re-training a model from scratch.
Therefore, the accumulated learning time is a rational metric 
for efficiency comparison in this context.
}, 
i.e. the summed time for training all the IRS models when 
the label number is increased from $50$ to $200$ one by one (in total $151$ updates),
we interpolated estimations on training time between these four
measured test cases.
\changed{A one-by-one model update is necessary particularly
when deploying a pre-trained sub-optimal re-id model to a previously unseen camera network
with weak starting performance.} 

We adopted the LOMO visual feature on all datasets. 
We utilised the 485/486 people split on CUHK01, 
the manually labelled person images on CUHK03,
the single-query setting on Market-1501,
and the same test data as the experiments in Sec.\ref{sec:eval_batch}. 
We conducted 10 folds of evaluations each with a different set of random unlabelled identities
and reported the averaged results.

\vspace{0.1cm}
\noindent \textbf{(I) Passive Incremental Learning. } 
We compared the proposed incremental learning (IL) based IRS (IRS$^\text{inc}$)
with the batch-wise learning (BL) based IRS in Table \ref{tab:incremental}
for model training time and re-id Rank-1 performance.
It is found that IRS model training speed can increase 
by one order of magnitude or more, with higher speed-up observed
on larger datasets and resulting in more model training efficiency gain. 
Specifically, on VIPeR, BL took approximately $36.5$ seconds to
conduct the $151$ model updates by re-training, while IL only
required $3.28$ seconds. 
When evaluated on Market-1501, 
BL took over $5.5$ hours ($1.9\times10^4$ seconds) to perform the sequential model updates, 
while IL was more than $20\times$ faster, only took $877.3$ seconds. 
Importantly, this speed-up is at the cost of only $1\!\!\sim\!\!2$\% Rank-1 drop.
This suggests an attractive trade-off for the IRS$^\text{inc}$ algorithm
between effectiveness and efficiency in
incremental model learning.
\vspace{0.2cm}
\noindent \textbf{(II) Active Incremental Learning. } 
We further evaluated the effect of the proposed {\em Joint$\text{E}^2$} 
active learning algorithm (Sec. \ref{sec:active})
by random passive 
unlabelled image selection ({\em Random}).
%
Also, we compared with
a state-of-the-art density based active sampling method \citep{ebert2012ralf}
which prefers to query the densest region of unlabelled sample
space ({\em Density}).
For both active sampling methods, we used our IRS$^\text{inc}$ for re-id model training.
%
We evaluated the four test cases ($50,100,150,200$ labelled identities) as shown in Table \ref{tab:CMC_active}.

It is evident from Table \ref{tab:CMC_active} that:
{\bf (1)} On all four datasets, our Joint$\text{E}^2$ outperformed clearly
both {\em Random} and {\em Density} given varying numbers of labelled samples.
For example, when $50$ identities were labelled, 
the proposed Joint$\text{E}^2$ algorithm beats {\em Random} sampling
in Rank-1 by $4.0\%$($23.4\!-\!19.4$), $9.1\%$($29.9\!-\!20.8$),
$3.0\%$($25.1\!-\!22.1$), $9.0\%$($36.5\!-\!27.5$) on VIPeR, CUHK01,
CUHK03 and Market-1501, respectively.  
%
{\bf (2)} Our Joint$\text{E}^2$ model obtained similar or even better performance 
with less human labelling effort. 
For example, on Market-1501, by labelling $150$ identities, Joint$\text{E}^2$
achieved Rank-1 rate of $54.8\%$, 
surpassed Random ($54.3\%$) and Density ($53.9\%$) with a greater budget of $200$ identities. 

In summary, the results in Tables \ref{tab:incremental} and
\ref{tab:CMC_active} show clearly that 
the hybrid of our proposed IRS$^\text{inc}$ model and Joint$\text{E}^2$ active sampling method 
provides a highly scalable active incremental re-id model training framework, 
with attractive model learning capability and efficiency from less
labelling effort suited for real-world person re-id applications. 

\section{Conclusion}
\label{sec:conclusion}

In this work, we developed a novel approach to explicitly designing a
feature embedding space for supervised batch-wise and incremental person re-identification model
optimisation. We solved the re-id model learning problem by introducing an
identity regression method in an Identity Regression Space (IRS)
with an efficient closed-form solution. 
Furthermore, 
we formulated an incremental learning algorithm IRS$^\text{inc}$ to
explore sequential on-line labelling and model updating. 
This enables the model to not only update efficiently the re-id model
once new data annotations become available, 
but also allows probably early re-id deployment and 
improves adaptively the re-id model to new test domains with potential temporal dynamics. 
To better leverage human annotation effort, we further 
derived a novel active learning method JointE$^2$ to selectively query
the most informative unlabelled data on-line.
Extensive experiments on four benchmarks show that our IRS method outperforms 
existing state-of-the-art re-id methods in the conventional batch-wise
model learning setting.
Moreover, the proposed incremental learning algorithm increases significantly
model training speed, over $10$ times faster than batch-wise model learning, 
by only sacrificing marginal model re-id capability with
$1$$\sim$$2\%$ Rank-1 drop. 
This labelling-while-deploying strategy has the intrinsic potential of
helping reduce the cost of manual labelling in large scale deployments by 
structuring semantically the unlabelled data so to expedite 
the true match identification process. 
Additionally, our active learning method improves 
notably the human labelling quality w.r.t. the thus-far model, particularly
when limited budget is accessible, providing over $3\%$ Rank-1
improvement than Random sampling given $50$ identities labelling
budget. 
While person re-id has attracted increasing amount of efforts 
especially in the deep learning paradigm,
model learning scalability, model incremental adaptation, and labelling effort minimisation in
large scale deployments however are significantly underestimated although very critical in real-world applications.
By presenting timely an effective solution in this work,
we hope that more investigations towards these important problems
will be made in the future studies.
One interesting future direction is to develop incremental deep re-id
learning algorithms.

\section*{Acknowledgments}

This work was partially supported by the China Scholarship Council, 
Vision Semantics Ltd, Royal Society Newton Advanced Fellowship Programme (NA150459),
and Innovate UK Industrial Challenge Project on Developing and Commercialising Intelligent Video Analytics Solutions for Public Safety (98111-571149).

\begin{appendix}

\section{Derivation of FDA Coding}
\label{sec:app}
In the following, we provide a detailed derivation of FDA coding (Eq. \eqref{eq:FDA_Y}) in our IRS method.
%
%

\vspace{0.2cm}
\noindent{\em FDA Criterion. } 
Specifically, the FDA criterion aims to minimise the intra-class (person) appearance variance and maximise inter-class appearance variance. 
Formally, given zero-centred training data $\bm{X} = \{\bm{x}_i\}_{i=1}^n$, 
we generate three scatter matrices defined as follows:
\begin{align}
\begin{split}
\bm{S}_w &= \frac{1}{n}\sum_{j=1}^c \sum_{l_i=j} (\bm{x}_i - \bm{u}_j)(\bm{x}_i - \bm{u}_j)^\top, \\
\bm{S}_b &= \frac{1}{n}\sum_{j=1}^c n_j \bm{u}_j \bm{u}_j^\top,\\
\bm{S}_t &= \bm{S}_w + \bm{S}_b= \frac{1}{n}\sum_{i=1}^n \bm{x}_i \bm{x}_i^\top,
\end{split}
\label{eq:scatter}
\end{align}
where $\bm{S}_w$, $\bm{S}_b$, and $\bm{S}_t$ denote {\it within-class}, {\it between-class} and {\it total}
scatter matrices respectively, $\bm{u}_j$ the class-wise centroids,
and $n_j$ the sample size of the $j$-th class (or person). 
The objective function of FDA aims at maximising $trace(\bm{S}_b)$ and minimising $trace(\bm{S}_w)$ simultaneously, where $\bm{S}_w$ can be replaced by $\bm{S}_t$ 
since $\bm{S}_t = \bm{S}_b + \bm{S}_w$. 
Hence, an optimal transformation $\bm{G}^*$ by FDA can be computed by solving the following problem:
\begin{align}
\bm{G}^* =  \arg\max_{\bm{G}} \; {trace} \Big(\big(\bm{G}^\top\bm{S}_b\bm{G}\big)\big(\bm{G}^\top\bm{S}_t\bm{G}\big)^\dagger\Big).
\label{eq:FDA}
\end{align}

\vspace{0.2cm}
\noindent {\it \textbf{Theorem 1.} 
	With $\bm{Y}$ defined as Eq. \eqref{eq:FDA_Y}, the projection $\bm{P}^*$ learned by Eq. \eqref{eq:HER_solution} is equivalent to $\bm{G}^*$, the optimal FDA solution in Eq. \eqref{eq:FDA}.}

\vspace{0.2cm}
\noindent \textbf{\em Proof.} 
First, optimising the objective in Eq. \eqref{eq:FDA_Y} involves solving the following eigen-problem:
\begin{equation}
\bm{S}_t^\dagger\bm{S}_b\bm{G} = \bm{G}\bm{\Lambda},
\label{eq:eigen_original}
\end{equation}
where $\bm{G} \in \mathbb{R}^{d\times q} = \left[\bm{g}_1, \cdots, \bm{g}_q\right]$ contains $q$  eigenvectors of $\bm{S}_t^\dagger\bm{S}_b$, and $\bm{\Lambda} = diag(\alpha_1, \cdots, \alpha_q)$  with $\alpha_i$ the corresponding eigenvalue, and $q = rank(\bm{S}_b) \leq c - 1$. 
From the definitions in 
Eq. \eqref{eq:scatter} and Eq. \eqref{eq:FDA_Y}, 
$\bm{S}_t$ and $\bm{S}_b$ can be further expanded as:
\begin{align}
\bm{S}_t &= \bm{X}\bm{X}^\top, \quad
\bm{S}_b = \bm{X}\bm{Y}\bm{Y}^\top\bm{X}^\top.
\end{align}
Here, the multiplier $\frac{1}{n}$ is omitted in both scatter matrices for simplicity.
Now, we can rewrite the left-hand side of Eq. \eqref{eq:eigen_original} as:
\begin{equation}
(\bm{X}\bm{X}^\top + \lambda\bm{I})^\dagger\bm{X}\bm{Y}\bm{Y}^\top\bm{X}^\top\bm{G} = \bm{G}\bm{\Lambda}.
\label{eq:eigen_expand}
\end{equation}
Note that, the pseudo-inverse $\bm{S}_t^\dagger$ is calculated by $(\bm{X}\bm{X}^\top + \lambda\bm{I})^\dagger$. 
The reason is that in real-world problems such as person re-id 
where training data is often less sufficient, 
$\bm{S}_t$ is likely to be ill-conditioned, i.e.
singular or close to singular, so that its inverse cannot be accurately computed.

By our solution $\bm{P}$ in Eq. \eqref{eq:HER_solution}, 
we can further rewrite Eq. \eqref{eq:eigen_expand}:
\begin{equation}
\bm{P}\bm{Y}^\top\bm{X}^\top\bm{G} = \bm{G}\bm{\Lambda}
\end{equation}  

To connect the regression solution $\bm{P}$ and the FDA solution $\bm{G}$, we define
a $c\times c$ matrix $\bm{R} = \bm{Y}^\top\bm{X}^\top\bm{P}$. 
According to the general property of eigenvalues~\citep{horn2012matrix}, 
$\bm{R}$ and $\bm{P}\bm{Y}^\top\bm{X}^\top$ share the same $q$ non-zero eigenvalues. Also, if  $\bm{V} \in \mathbb{R}^{c\times q}$ contains the $q$ eigenvectors of $\bm{R}$, columns of the matrix $\bm{P}\bm{V}$ must be the eigenvectors of the matrix $\bm{P}\bm{Y}^\top\bm{X}^\top$. 
Therefore, the relation between $\bm{P}$ and $\bm{G}$ is:
\begin{equation}
\bm{G} = \bm{P}\bm{V}
\label{eq:GP}
\end{equation}

Finally, we show in the following Lemma  that $\bm{P}$ and $\bm{G}$  are equivalent in the aspect of re-id matching.

\vspace{0.2cm}
\noindent {\it \textbf{Lemma 1.} In the embedding provided by $\bm{P}$ and $\bm{G}$, the nearest neighbour algorithm produce same result. That is, $(\bm{x}_i - \bm{x}_j)^\top \bm{P}\bm{P}^\top(\bm{x}_i - \bm{x}_j) = (\bm{x}_i - \bm{x}_j)^\top \bm{G}\bm{G}^\top(\bm{x}_i - \bm{x}_j)$}.

\vspace{0.2cm}
\noindent \textbf{Proof.} The necessary and sufficient condition for  Lemma 1 is $\bm{P}\bm{P}^\top = \bm{G}\bm{G}^\top$. As $\bm{V} \in \mathbb{R}^{c\times q}$, there must exist a matrix $\bm{V}_2 \in \mathbb{R}^{c\times (c-q)}$ such that
$\bm{\hat{V}} = [\bm{V}, \bm{V}_2]$ is a $c\times c$ orthogonal matrix. 
Suppose the diagonal matrix $\bm{\Gamma}$ contains the non-zero eigenvalues of $\bm{R}$, then the eigen decomposition $\bm{R}=\bm{V}\bm{\Gamma}\bm{V}^\top$ implies that  $\bm{V}_2^\top\bm{R}\bm{V}_2 = 0$. 

Recall that $\bm{R} = \bm{Y}^\top\bm{X}^\top\bm{P}$, and $\bm{P} = (\bm{X}\bm{X}^\top + \lambda \bm{I})^\dagger\bm{X}\bm{Y}$, then we obtain: 
\begin{equation}
\bm{V}_2^\top \bm{Y}^\top \bm{X}^\top (\bm{X}\bm{X}^\top + \lambda \bm{I})^\dagger\bm{X}\bm{Y} \bm{V}_2 = 0
\end{equation}
As $(\bm{X}\bm{X}^\top + \lambda \bm{I})^\dagger$ is positive definite, the above equation implies that $\bm{X}\bm{Y}\bm{V}_2 = 0$, and hence $\bm{P}\bm{V}_2 =(\bm{X}\bm{X}^\top + \lambda \bm{I})^\dagger\bm{X}\bm{Y} \bm{V}_2 = 0$. Hence, we have:
\begin{align}
\begin{split}
\bm{P}\bm{P}^\top &= \bm{P}\bm{\hat{V}}\bm{\hat{V}}^\top \bm{P}^\top \\
&= \bm{P}\bm{V}\bm{V}^\top \bm{P}^\top + \bm{P}\bm{V}_2\bm{V}_2^\top \bm{P}^\top \\
&= \bm{G}\bm{G}^\top + 0 
\end{split}
\end{align}
As such, the proof to Lemma 1 and Theorem 1 is complete.

\end{appendix}


\bibliographystyle{spbasic}      

\bibliography{IJCV17_IRS_CR.bbl}

\end{document}